\documentclass[times,twocolumn,final]{elsarticle}

\usepackage{medima}
\usepackage{framed,multirow}

\usepackage[authoryear]{natbib}
\usepackage{booktabs}
\usepackage{arydshln}
\usepackage{threeparttable}
\usepackage{color,graphicx}
\usepackage{amsmath}
\usepackage{mathrsfs}
\usepackage{amsthm}
\newtheorem{assumption}{Assumption}
\usepackage[ruled]{algorithm2e}
\usepackage{bm}
\usepackage{multirow}
\usepackage{amssymb}
\usepackage{xcolor}

\newcommand{\ie}{\textit{i.e.,} } 
\providecommand{\zxhreftb}[1]{Table~\ref{#1}} 
\providecommand{\zxhreffig}[1]{Figure~\ref{#1}} 
\usepackage[commandnameprefix=always]{changes}
\usepackage{subfigure}
\usepackage[switch]{lineno}
\definecolor{navy}{RGB}{9,123,248}
\definecolor{red}{RGB}{215,59,46}

\usepackage{verbatim}

\usepackage{latexsym}

\usepackage{url}
\usepackage{xcolor}

\definecolor{newcolor}{rgb}{.8,.349,.1}

\journal{Medical Image Analysis}

\usepackage{hyperref}

\begin{document}

\verso{K.Zhang \textit{et~al.}}

\begin{frontmatter}

\title{ZScribbleSeg: A comprehensive segmentation framework with modeling of efficient annotation and maximization of scribble supervision}

\author[1,2]{Ke \snm{Zhang}\fnref{equal}}
\author[1]{Bomin \snm{Wang}\fnref{equal}}
\author[1]{Hangqi \snm{Zhou}}
\fntext[equal]{These authors contributed equally to this work.}

\author[1]{Xiahai \snm{Zhuang}\corref{*}}
\cortext[*]{Xiahai Zhuang is the corresponding author. Email: zxh@fudan.edu.cn; Ke Zhang currently is a PhD student at Johns Hopkins University, and this work was done when she was in Fudan University.}

\address[1]{School of Data Science, Fudan University, Shanghai, 200433, China}
\address[2]{Department of Electrical and Computer Engineering, Johns Hopkins University, Baltimore, USA}


\begin{abstract}
Curating fully annotated datasets for medical image segmentation is labour-intensive and expertise-demanding. To alleviate this problem, prior studies have explored scribble annotations for weakly supervised segmentation. Existing solutions mainly compute losses on annotated areas and generate pseudo labels by propagating annotations to adjacent regions. However, these methods often suffer from inaccurate and unrealistic segmentations due to insufficient supervision and incomplete shape information.
In contrast, we first investigate the principle of good scribble annotations, which leads to efficient scribble forms via supervision maximization and randomness simulation. We further introduce regularization terms to encode the spatial relationship and the shape constraints, where the EM algorithm is utilized to estimate the mixture ratios of label classes. These ratios are critical in identifying the unlabeled pixels for each class and correcting erroneous predictions, thus the accurate estimation lays the foundation for the incorporation of spatial prior.
Finally, we integrate the efficient scribble supervision with the prior into a framework, referred to as ZScribbleSeg, and apply it to multiple scenarios. Leveraging only scribble annotations, ZScribbleSeg achieves competitive performance on six segmentation tasks including ACDC, MSCMRseg, BTCV, MyoPS, Decathlon-BrainTumor and Decathlon-Prostate. Our code will be released via \href{https://github.com/DLwbm123/ZScribbleSeg}{\textcolor{black}{https://github.com/DLwbm123/ZScribbleSeg}}.
\end{abstract}


\end{frontmatter}

\section{Introduction}\label{sec:introduction}
In recent years, deep neural networks has demonstrated its potential on various visual tasks~\citep{zhao2024review}.  
However, the success of these methods relies on massive annotations, which require labor-intensive manual efforts.
For medical imaging, the dense manual labeling can take several hours to annotate just one image for experienced doctors, which is both expensive and expertise-demanding~\citep{zhuang2019multivariate}.
Numerous efforts have contributed to the area of training segmentation networks with weaker annotations, including scribbles~\citep{han2024dmsps}, bounding boxes~\citep{wei2023weakpolyp}, points~\citep{lin2023nuclei}, and image-level labels~\citep{kuang2023cluster}.
Numerous studies have been reported utilizing only image-level labels~\citep{kuang2023cluster,zhang2021affinity,wang2022looking}. These methods mainly rely on large-scale training datasets, and tend to underperform on small medical image datasets.
On the contrary, scribbles are suitable for labeling nested structures and easy to obtain in practice. 
Several works have demonstrated their potential on both semantic and medical image segmentation~\citep{han2024dmsps,liu2022weakly,khoreva2017simple}.
Therefore, we aim to investigate this specific form of weakly supervised medical image segmentation using only scribble annotations for model training.
	
Conventionally, scribble annotations are mainly focused on delineating the structures of interest~\citep{9389796}. 
This can be effective in segmenting \textit{regular structures} with fixed shape patterns. 
Hence, this task is often referred to as \textit{regular structure segmentation}. 
However, such methods are challenged when applied to irregular targets with heterogeneous distributions, such as pathologies. 
This is also referred to as \textit{irregular (object) segmentation}, which is particularly challenging for the medical tasks with small training datasets.
Existing scribble learning approaches mainly typically focus on reconstructing complete labels from scribbles and then using the resulting pseudo labels for training.
These methods include \textbf{1)} label expansion strategies that assume the pixels with similar features are likely to be in the same category~\citep{ji2019scribble}, and \textbf{2)} ensemble methods that generate labels by fusing several independent predictions~\citep{luo2022scribble}.
These methods are prone to label noise from inaccurate segmentation proposals.
To overcome this issue, ~\cite{obukhov2019gated} proposed a regularization loss, which exploited the similarity between labeled and unlabeled regions.
Adversarial learning approach has also been applied to scribble-supervised segmentation by leveraging shape prior provided by additional full annotations~\citep{9389796}.
    
Scribble-supervised segmentation generally suffers from inadequate supervision and imbalanced label classes.
This often results in \textit{under segmentation} of target structures, where the volumes of segmented structures tend to be shrunk (see Section~\ref{problems}).
To address the problem of inadequate supervision, we first investigate the principles of generating \textit{good scribbles}, which serve both as guidance for annotation design and as a foundation for supervision augmentation.  
The goal is to design efficient scribbles that maximize supervision without additional annotation costs.
Our studies demonstrate that the model training benefit from the randomness of wide range distributed scribbles and larger proportion of annotated areas.
Inspired by this, we propose to simulate such types of scribble-annotated images as a means of \textit{supervision augmentation}. 
This can be achieved via mixup and occlusion operations on existing training images, and the supervision augmentation is coupled with regularization terms penalizing any inconsistency in the segmentation results.  

Despite the lack of supervision, the scribble annotations typically have imbalanced annotated label proportions, leading to biased shape information.
As a result, the model cannot accurately capture the global shape of target structures.
We therefore further propose to correct the problematic prediction using prior-based regularization. 
This requires the critical step of estimating the mixture proportion (ratio) of each label class (referred to as  $\bm\pi$ prior).
We propose an algorithm to estimate this prior and use it to construct a spatial loss defined on the marginal probabilities of pixel labels and spatial energy.
This spatial loss is a regularization term aimed to correct the shape of segmentation results.
The supervision augmentation and prior-based regularization work in a complementary way, and both contribute to the stable and robust training on a variety of segmentation tasks.

The proposed scribble-supervised segmentation method, referred to as ZScribbleSeg, extends and generalizes the algorithms in our two preliminary works~\citep{zhang2022cyclemix,zhang2022shapepu}, and has more scientific  significance in the following aspects:
Firstly, we investigate principles of efficient scribble forms to guide the supervision augmentation, which have never been reported to the best of our knowledge.
Secondly, we propose a new strategy leveraging spatial prior to adjust the predicted probability with computed spatial energy.
Thirdly, we generalize ZScribbleSeg to the 3D cases, and validate the proposed framework on five use cases (datasets), including 3D Prostate segmentation (Decathlon-Prostate), cardiac structural segmentation (ACDC), LGE segmentation (MSCMRseg), and myocardial pathology segmentation combining three-sequence MRI (MyoPS).

The contributions of this paper are summarized as follows.
    
\begin{itemize}

\item We propose a comprehensive framework for scribble-supervised segmentation that models efficient scribbles and corrects network predictions with prior regularization. This framework aims to alleviate the problems of inadequate supervision and under-segmentation.
\item To augment the supervision of scribble annotations, we first investigate the principles of efficient scribble forms.
Motivated by the findings that network benefits from larger and randomly distributed annotation, we model efficient scribbles by maximizing supervision via mixup and enhancing randomness via occlusion.
\item To tackle the unders-egmentation and shape distoration problems, we propose a prior-guided mechanism to correct the shape of model prediction based on prior regularization, including $\bm{\pi}$ prior, spatial prior, and shape constraints.
We utilize the EM algorithm to estimate $\bm{\pi}$ prior, based on which we propose the spatial prior loss to encode spatial relationship.
\item Our approach achieved competitive performance in weakly supervised segmentation across diverse tasks, including regular structures in cardiac anatomy, regular structures in pathology-enhanced imaging, irregular pathological objects, and 3D prostate segmentation.
\end{itemize}
\begin{figure}[!t]
    \centering
    \includegraphics[width=0.45\textwidth]{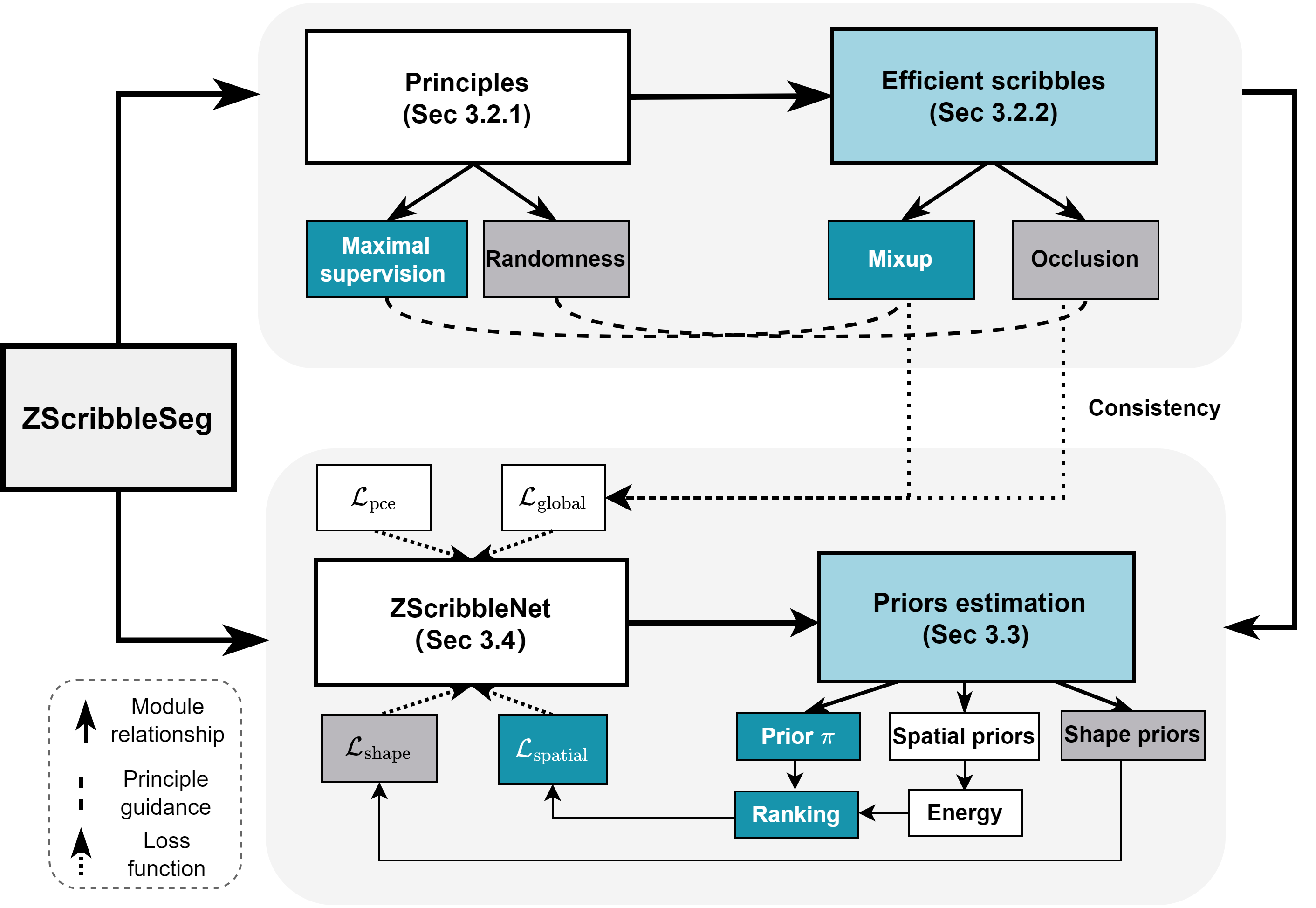}
\caption{Roadmap of the proposed ZScribbleSeg framework.}
\label{fig:roadmap}
\end{figure}

The rest of this paper is organized as follows: Section 2 briefly introduces the relevant research. 
In Section 3, we describe the modelling of efficient scribbles and the computation of the prior.
Section 4 presents the results of the efficiency, ablation, and validation study.
Finally, we conclude this work in Section 5.

\section{Related work}
This section provides a brief review of weakly supervised segmentation methods. Besides, we describe data augmentation strategies and regularization loss functions that are closely related to our work.
	
\subsection{Weakly supervised segmentation}
Recently, a variety of weakly supervised segmentation strategies have been developed to reduce the manual annotation efforts~\citep{lin2024fedlppa, yang2024anomaly, kweon2023weakly, liu2025weakly}. 
Among them, the scribbles are of particular interest for the application to medical image annotation, given by their advantage in annotating nested structures.
Current weakly supervised learning methods with image-level annotations mainly generate label seeds with Class Activation Map (CAM)~\citep{zhou2016learning} at first, and then train the network with refined pseudo labels.
However, the training of CAM requires a large scale of training data labeled with rich visual classes, which is not practical in clinical applications. 
Therefore, we investigate the scribble-supervised segmentation, due to its efficiency and effectiveness in both medical scenarios.

Scribble is a form of sparse annotation that provides labels for a small subset of pixels in an image~\citep{tajbakhsh2020embracing}.
Previous approaches mainly calculate losses for annotated pixels.
One group of works is designed to expand the annotations and reconstruct the full label for network training.
However, the expansion of labels needs to be achieved through iterative computation, which is particularly time-consuming.
To alleviate it, several works removed the relabeling process and instead adopted conditional random fields to perform the refinement of segmentation results \cite{Can2018LearningTS,Tang2018OnRL}.
However, the common issue is the unstable model training caused by noisy pseudo labels. 

To obtain high-quality pseudo labels and update it throughout the training process, ~\cite{luo2022scribble} proposed to mix the predictions from a dual-branch network as auxiliary pseudo labels.
This approach has achieved promising results on cardiac segmentation, but is still susceptible to inaccurate supervisions, especially on more challenging tasks with irregular objects.
ScribFormer ~\cite{li2024scribformer} introduces a transformer-based approach for scribble-supervised medical image segmentation. It adopts a hybrid CNN–Transformer design, where CNN modules capture fine-grained local spatial details and Transformer blocks model long-range contextual dependencies.
~\cite{chentip25} addresses inconsistent and incomplete scribble annotations by constructing a reference set from class-specific tokens and pixel-level features and using it as a shape prior to guide pixel-level feature matching for unlabelled pixels.
HELPNet ~\cite{zhangmedia25} introduces a hierarchical perturbation consistency module to improve feature learning via density-controlled jigsaw perturbations across global, local, and focal views, enabling multiscale structural representation learning. It further uses an entropy-guided pseudo-label module to estimate prediction confidence and generate pseudo-labels, and a structural prior refinement module that incorporates connectivity analysis and image boundary priors to refine pseudo-label quality.

Other works~\cite{9389796,zhang2020accl} included new modules to evaluate the quality of segmentation masks, which encourages the predictions to be realistic, but requiring extra full annotations.~\cite{gao2022segmentation} introduced a weakly- and semi-supervised segmentation framework that learns from sparse scribble annotations and large unlabeled datasets using teacher-student consistency and a multi-angle projection reconstruction loss. It demonstrates strong robustness and generalization on multiple datasets.~\cite{effdnetmiccai25} combines a foreground-background separation loss and a foreground augmentation strategy with diverse context to improve foreground discrimination and generalization.~\cite{zhoumiccai23} combines superpixel-guided scribble walking for denser supervision expansion and class-wise contrastive regularization for improving class feature compactness under scribble supervision.~\cite{zhoumva25} introduced Scribble2D5, a scribble-supervised 2.5D framework for volumetric segmentation that combines label propagation, boundary prediction, and an optional shape-prior module using unpaired segmentation masks.
Scribbles have been also adopted for user interaction to improve the results predicted by a model.
A group of studies~\citep{asad2022econet,roth2021going} investigated the interactive segmentation methods based on scribbles, to refine results, accelerate training and facilitate fine-tuning.

\subsection{Data augmentation}
Augmentation methods are investigated to improve the model generalization ability, by synthesizing virtual training examples in the vicinity of the training dataset~\citep{garcea2023data}.
Common strategies include random cropping, rotation, flipping and adding noise.
A line of research works have been proposed on Mixup augmentation~\citep{qin2024sumix, zou2023benefits, zhang2018mixup,kimICML20,kim2021comixup}, which blends two image-label pairs to generate new samples for classification tasks.
Input Mixup~\citep{zhang2018mixup} was introduced to perform linear interpolation between two images and their labels. 
~\cite{kimICML20} proposed Puzzle Mix to leverage the saliency and local statistics to facilitate image combination.
SUMix~\cite{qin2024sumix} futher proposed to learn the mixing ratio as well as the uncertainty for the mixed samples during the training process

For medical image analysis, Mixup methods have been adopted for image segmentation~\citep{chaitanya2019semi} and object detection tasks~\citep{wang2020focalmix}.
Although mixup operation may generate unrealistic samples, mixed soft labels can provide rich information and improve the model performance on semi-supervised segmentation~\citep{chaitanya2019semi}.

\subsection{Regularization losses}
Neural networks perform pixel-wise image segmentation, typically trained with cross-entropy or Dice loss.
To predict coherent segmentation in the global sense~\citep{kohl2018probabilistic}, several methods are proposed to regularize the neural network training.
Here, we focus on the consistency regularization and prior regularization that most relevant to our work. In addition, Perturbation-based interpretability methods (e.g., Meaningful Perturbations~\citep{Fong_iccv17} and Extremal Perturbations~\citep{Fong_iccv19}) optimize a perturbation mask under sparsity constraints to identify evidence regions for post-hoc classifier explanations. In contrast, our method does not optimize a perturbation mask. Instead, during segmentation training, we rank a class-wise spatial energy map, use the estimated class mixture ratio ($\pi$ prior) to select class-specific pixel subsets, and apply this process as a regularizer for prediction correction in scribble-supervised segmentation.

The consistency regularization leverages the fact that the perturbed versions of the same image patch should have the consistent segmentation.
A series of research has been conducted on consistency regularization~\citep{zhu2017unpaired,ouali2020semi}. Regularization losses have also been used in segmentation to suppress false positives and constrain prediction extent. For example, uncertainty-based background regularization can be implemented by maximizing prediction entropy over background regions, together with region-balancing constraints to avoid trivial solutionsp~\cite{Belharbitmi21}. In addition,~\cite{Kervadecmedia19, jiatmi17} regularize predictions by imposing externally specified region-size constraints (e.g., target size ranges or expert-provided area estimates). In contrast, our prior regularization is designed to address class imbalance and under-segmentation by first estimating class mixture ratios (the $\pi$ prior) on unlabeled pixels and then using this estimated prior to construct spatial regularization for prediction correction.
    
The proposed regularization of $\bm{\pi}$ prior is inspired from the binary mixture proportion estimation~\citep{bekker2018estimating,garg2021mixture,ramaswamy2016mixture}, which was originally designed for binary (two-class) positive unlabeled learning~\citep{du2015convex,NIPS2017_7cce53cf}.
For multi-class segmentation, the mixture ratios of classes are both imbalanced and inter-dependent, which cannot be solved by existing binary estimation methods.
	        
 \begin{figure*}
    \centering
    \includegraphics[width=0.9\textwidth]{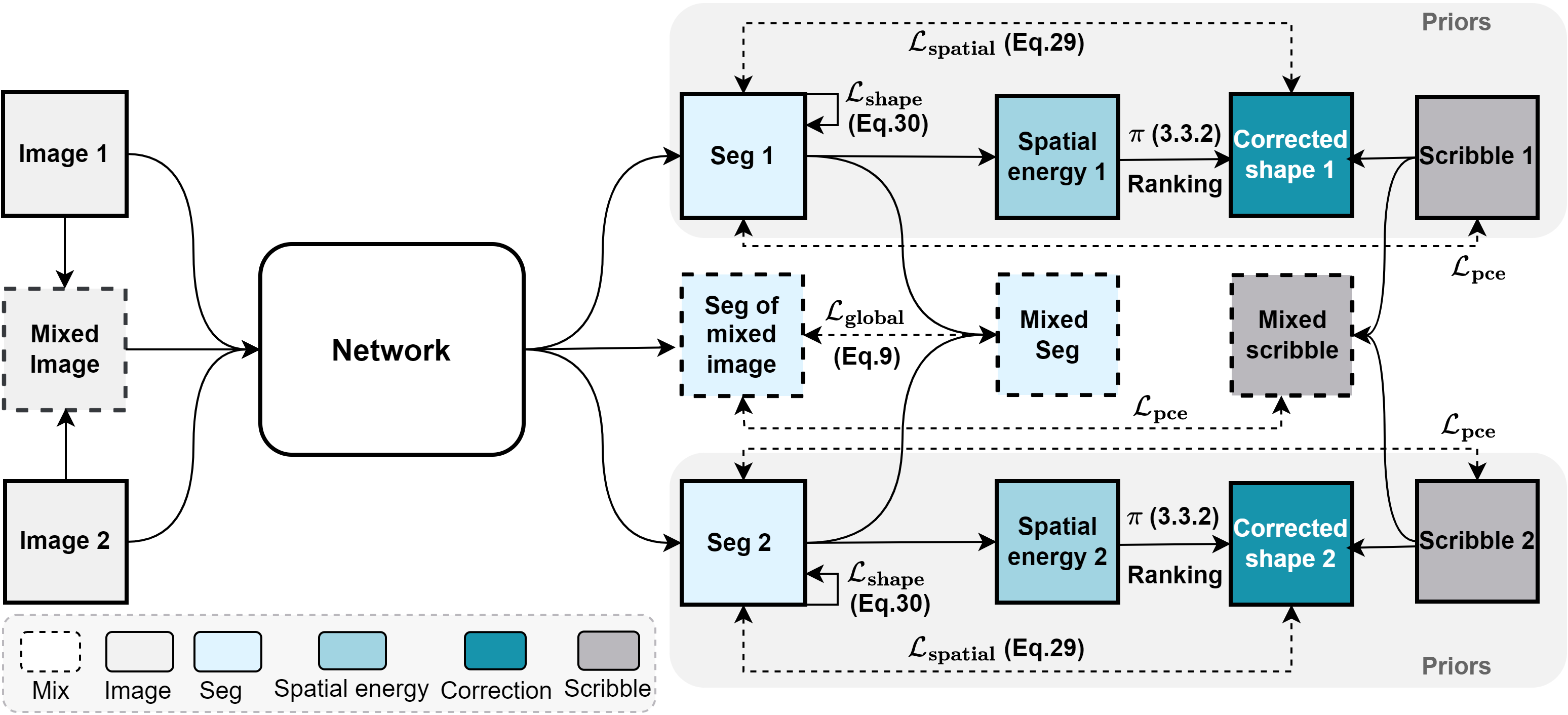}
\caption{
Overview of the training losses for the proposed ZScribbleNet, which consists of modeling of efficient scribbles, computation of priors and shape regularization term. 
The scribble modeling includes mixup augmentation, regularized with global consistency ($\mathcal{L}_{\text{global}}$). 
The prior comprises class mixture ratios ($\bm{\pi}$) and  spatial prior. These components contribute to the spatial prior loss ($\mathcal{L}_{\text{spatial}}$) and the shape regularization loss} ($\mathcal{L}_{\text{shape}}$).
Note that spatial prior loss is complementary with the partial cross entropy loss ($\mathcal{L}_{\text{pce}}$) which is solely calculated for labeled pixels.

\label{fig:overview}
\end{figure*}

\section{Method}
\subsection{Overview}
\noindent\newline\textbf{Problem Setup:} 
This work investigates the scenario of scribble-supervised segmentation, where the training images are solely annotated with a small number of pixels, via scribbles for each label class.

\noindent\newline\textbf{Motivation and Strategy:} 
To augment the supervision of scribble annotations, we first investigate different forms of scribbles. This allows us to establish principles of efficiency that maximize supervision without additional scribble effort.
These principles enable effective and robust model training with minimal annotation cost. Motivated by the principles, we introduce supervision augmentation and consistency regularization to maximize supervision while preserving global consistency. To tackle the under-segmentation problem, we further propose a prior-guided regularization scheme to exploit shape information and correct network predictions. Specifically, we first utilize the EM algorithm to estimate the class mixture ratios. Building on these estimates, we introduce a spatial prior loss by ranking spatial energy and selecting class-specific pixels. In addition, because scribble annotations are sparse, the network often produces fragmented predictions with multiple disconnected components. We therefore introduce a shape regularization term to promote inter-connectivity within the structure.

\noindent\newline\textbf{Solution:}
We develop ZScribbleSeg consisting of
(1) modeling efficient scribbles via supervision maximization and randomness simulation;
(2) modeling and computation of the priors, including label class proportion prior, spatial prior, and shape constraints;
(3) integration to develop deep neural network (referred to as ZScribbleNet) having losses of partial cross entropy ($\mathcal{L}_{\text{pce}}$), global consistency $(\mathcal{L}_{\text{global}})$, spatial prior loss ($\mathcal{L}_{\text{spatial}}$), shape regularization $(\mathcal{L}_{\text{shape}})$ and training strategy of supervision augmentation and prior regularization. \zxhreffig{fig:roadmap} presents the roadmap of the proposed framework.

\subsection{Principle and modeling of efficient scribbles}
We investigate the principles of efficient scribbles and derive the objective of maximizing supervision with minimal annotation efforts.
This leads to the proposal of supervision augmentation. In addition, we propose a global consistency loss to penalize the non-equivalence in the augmentation.

\subsubsection{Principles of efficient scribbles}
We shall verify the two principles of achieving efficient scribble annotation in terms of maximal supervision later through the experiments in Section~\ref{sec-exp-scribble}:
\noindent\newline
(1) The large proportion of pixels annotated by scribbles compared with the whole set. 
When the amount of annotated pixels increases, the performance of the model will gradually converge to the upper bound of full annotations. Our goal is to investigate the effective strategies to compensate for missing annotations and match the performance of full annotations.

\noindent\newline
(2) The randomness of the distribution of scribbles. This is represented by the random and wide-range annotations. 

Firstly, we are motivated by the knowledge that model training benefits from the finer gradient flow through a larger proportion of annotated pixels~\citep{tajbakhsh2020embracing}.
Therefore, we try to increase the annotation proportion with same effort. 
One natural idea is to simply expand the width of scribbles.
However, this way only increases the label amount in the local area, and lacks the ability to enlarge the annotation range across the entire image.
    
Secondly, we are inspired by the fact that the imaging data are easier to be restored from random samples of pixels than from down-sampled low-resolution images with regular patterns~\citep{gao2020robust}.
This was due to the fact that the randomly and sparsely distributed samples maintain the global structure of the imaging data, which therefore can be restored with existing low-rank or self-similarity regularization terms. 
By contrast, the regularly down-sampled low-resolution images have evidently reduced tensor ranks, compared with the original high-resolution data, thus lose the global structure information. 
Motivated by this, we assume the features of full segmentation (similarly to the global structure information) can be portrayed (restored) with sparse scribble annotations randomly and widely distributed within the entire dataset. 
With such scribble annotation, the segmentation network can easily learn the global shape prior.
     
    
\subsubsection{Modeling via supervision augmentation}
Based on the observations that large-proportion annotated pixels and randomness of distribution lead to efficient scribbles, we propose to model efficient scribbles by supervision augmentation simulating large annotation proportion and randomness of scribble distribution. Specifically, we aim to generate training images with efficient scribbles by maximizing the supervision via mixup operations and achieving randomness via occlusion operations. To mightigate the shape distortion induced by mixup operations, we further propose a global consistency loss.
This resembles data augmentation, which increases the data diversity and enables robust training.

\noindent\newline\textbf{Search optimal annotation with mixup:} 
Motivated by the principles of efficient scribble, we first seek the optimal scribble with a large annotated ratio, high supervision, and unchanged local features.
To achieve that, instead of maximizing the annotations directly, \emph{we aim to maximize the saliency of mixed images}, which measures the sensitivity of the model to inputs. Specifically, We utilize PuzzleMix ~\citep{kimICML20} as the mixup operation.
Given that the annotated area tends to be accompanied by high saliency, maximizing saliency also increases the scribble annotations.

{We define an image–scribble pair $(X, Y)$ as an image $X$ together with its 
corresponding scribble annotation $Y$.  
We denote $(X_1, Y_1)$ and $(X_2, Y_2)$ are two different image-scribble pairs that are sampled 
independently from the training set with dimension $n$.} We denote the resulted mixed image-label pair as $(X'_{12},Y'_{12})$. 
The transportation process is defined by:
    \begin{gather*}
        X'_{12} = T(X_1, X_2)\ \text{\ and\ }\ Y'_{12} = T(Y_1, Y_2), \label{eq:miximage}\\
        T(X_1, X_2) = (1-\beta)\odot \textstyle\prod_1 X_1+ \beta\odot \textstyle\prod_2 X_2, \label{eq:mixop}
    \end{gather*}
where $T(X_1,X_2)$ represents the transportation process between image $X_1$ and $X_2$; 
$\scriptstyle\prod_i$ denotes the transportation matrix of size $n\times n$ for image $X_i$; $\beta$ means the mask with value $[0, 1]$ of dimension $n$; $\odot$ is the element-wise multiplication. 
Then, we aim to maximize the saliency of transportation result over the parameters $\{{\scriptstyle\prod_1}, {\scriptstyle\prod_2},\beta\}$:
    \begin{linenomath*}\begin{equation*}
    \{{\scriptstyle\prod_1}, {\scriptstyle\prod_2},\beta\} = 
    \underset{{\scriptstyle\prod_1},{\scriptstyle\prod_2}, \beta}{\arg\max}[ (1-\beta)\odot {\scriptstyle\prod_1} M(X_1) + \beta\odot {\scriptstyle\prod_2} M(X_2) ],
    \end{equation*}\end{linenomath*}
    
\begin{figure*}[!t]
    \centering
    \includegraphics[width=0.8\textwidth]{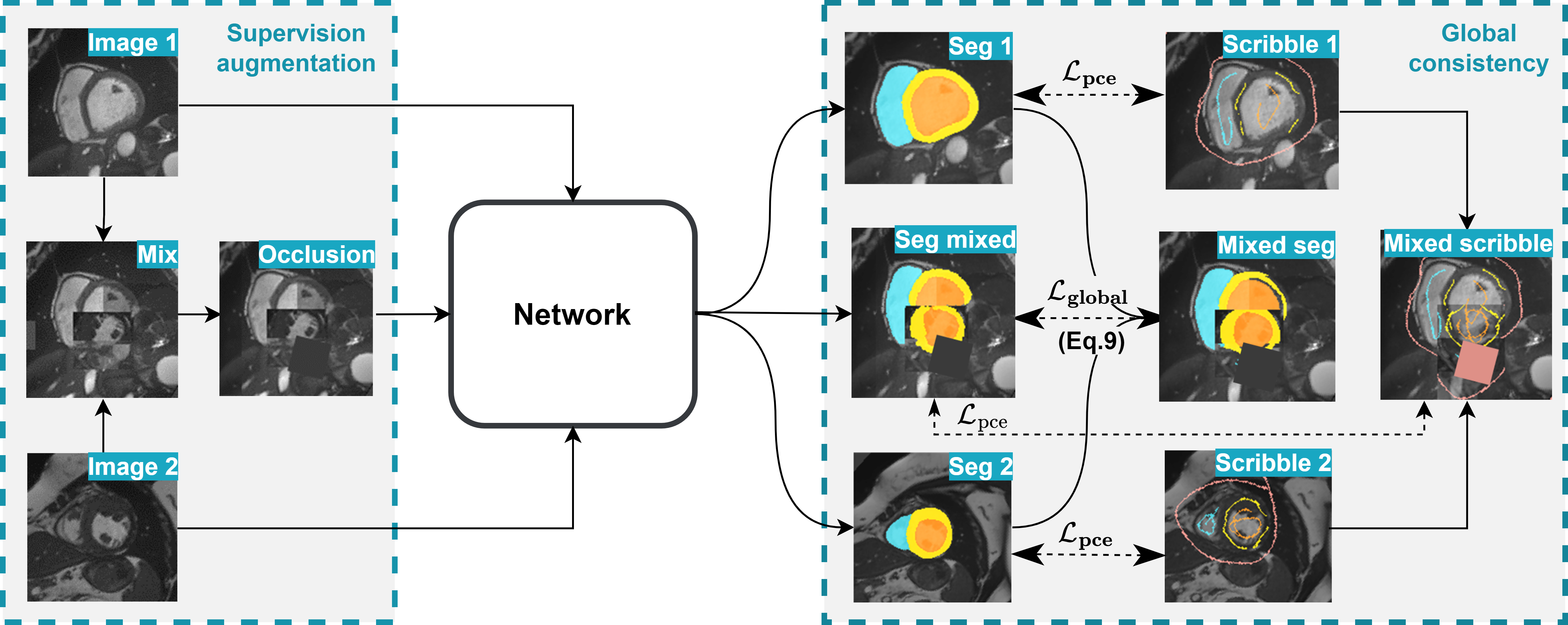}
\caption{Illustration of supervision augmentation and global consistency. Supervision maximization is achieved with the mix augmentation to increase the annotated proportion and data variety. Global consistency requires the segmentation result of mixed image and unmixed image to be consistent.
}
\label{fig:illustration}
\end{figure*}
where $M(X)$ denotes the saliency map of image $X$, which is obtained by computing the $l_2$ norm of gradient values. We solve this optimization problem based on PuzzleMix~\citep{kimICML20}. 
To preserve the local statistical features, the optimization objective also includes the image local smoothness and the mixing weight prior. 
For details of the optimization objective, we refer readers to PuzzleMix~\citep{kimICML20} and the Section 1 of the supplementary materials.

\noindent\newline\textbf{Introduce randomness via occlusion:} We propose to simulate randomly distributed scribbles via occlusion. Specifically, one square area of the mixed image is randomly dropped and its label is set to be the background.
Since that the proportion of the background annotated by scribbles tends to be smaller than that of the foreground classes, the occlusion operation alleviates the imbalance problem of class mixture ratios within labeled pixels, and further improves the results of mixture ratio estimation, which will be elaborated in Section~\ref{Sec.estimation}.
   
We denote the occluded image-label pair as $(X'', Y'')$, which is obtained by: 
    \begin{gather*}
        X''_{12} = (1-\mathbf{1}_b) \odot X'_{12}, \\
        Y''_{12} = (1-\mathbf{1}_b) \odot Y'_{12},
    \end{gather*}
where $\mathbf{1}_b$ denotes a rectangular mask of size $n\times n$ with value in $[0,1]$. 
The rectangular mask is randomly rotated to occlude the mixed image, and turns the occluded area into background.
Following~\citep{yun2019cutmix}, we set the size of rectangular to be $32\times32$. {This size is suitable for typical spatial scale of cardiac structures in 2D slices. Smaller masks introduce limited perturbation and fail to increase 
the diversity of labeled background pixels, whereas significantly larger masks remove too much image content and may disrupt local anatomical cues.}

\noindent\newline\textbf{Global consistency loss:}
    The objective of global consistency regularization is to leverage the mix-invariant property.
    As \zxhreffig{fig:illustration} shows, global consistency requires the same image patch to have consistent segmentation in two scenarios, \textit{i.e.}, the unmixed image and the mixed image.
    Let the segmentation result of image $X$ predicted by network be $\hat{Y} = f(X)$. For the transported image $X'_{12} = T(X_1,X_2)$, the consistency of mixup is formulated as:    
    \begin{linenomath*}\begin{equation*}
        T(f(X_1), f(X_2)) = f(T(X_1, X_2)),
    \label{3.1eq1}
    \end{equation*}\end{linenomath*}
    which requires the segmentation of mixed image to be consistent with the mixed segmentation, after the same transportation process. 
    When applying the occlusion operation, we further have:    
    \begin{linenomath*}\begin{equation}
        (1-\mathbf{1}_b)\odot T(\hat{Y}_1, \hat{Y}_2) 
        = 
        f\left((1-\mathbf{1}_b)\odot T(X_1, X_2)\right).
    \label{4.1eq2}
    \end{equation}\end{linenomath*}
    Then, we propose to minimize the distance between two sides of Eq.(\ref{4.1eq2}).
    Let $u_{12}=(1-\mathbf{1}_b)\odot T(\hat{Y}_1, \hat{Y}_2)$ and $v_{12} =f\left((1-\mathbf{1}_b)\odot T(X_1, X_2)\right)$. 
    The negative cosine similarity $\mathcal{L}_{n}(u_{12}, v_{12})$ is defined as:
    \begin{linenomath*}\begin{equation*}
        \mathcal{L}_{n}(u_{12}, v_{12}) = -\frac{u\cdot v}{||u_{12}||_2 \cdot ||v_{12}||_2}.
    \end{equation*}\end{linenomath*}    
    Taking the symmetrical metric into consideration, we similarly penalize the inconsistency between $u_{21}$ and $v_{21}$. 
    Therefore, the global consistency loss is formulated as:
    \begin{linenomath*}\begin{equation}
        \mathcal{L}_{\text{global}} = \frac{1}{2}\left[\mathcal{L}_{n}(u_{12},v_{12}) + \mathcal{L}_{n}(u_{21}, v_{21})\right].
    \label{4.1eq4}
    \end{equation}\end{linenomath*}        
\textbf{Discussion:} The mix operations of mixup and occlusion could change the shape of target structures, resulting in the unrealistic image. To tackle it, as \zxhreffig{fig:illustration} shows, we propose to combine the partial cross entropy (PCE) loss for labeled pixels of both mixed and unmixed images, and leverage mix equivalence to preserve shape consistency at the global level. To further exploit the shape features, we propose to correct the network prediction guided by the computed prior, which is described in Section~\ref{sec3.3}.
 
\begin{figure}[!t]
    \centering
    \includegraphics[width=\linewidth]{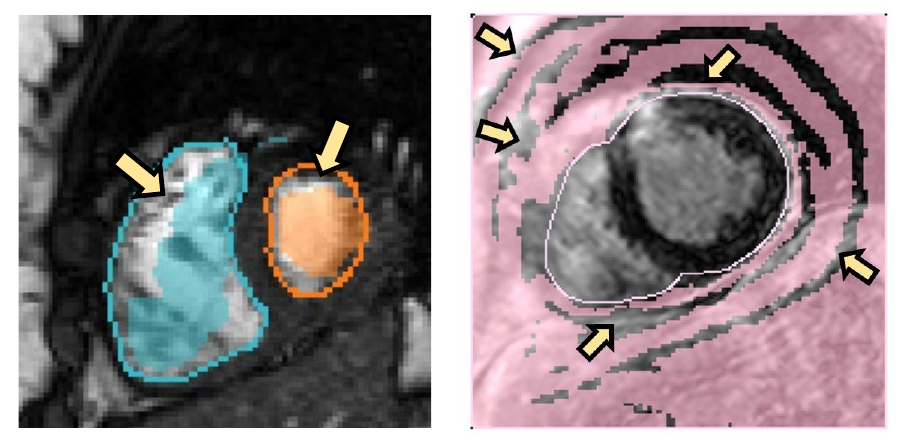}\\
    \makebox[\linewidth][s]{\ (a)\ \ (b)\ }
    \caption{Two under segmentation examples indicated by yellow arrows: (a) foreground classes (left and right ventricles) in ACDC dataset; (b) background in MyoPS dataset.
    } \label{fig:underseg}
\end{figure}

\subsection{Modeling and computation of the priors}
 \label{sec3.3}
\begin{figure*}[!t]
    \centering
    \includegraphics[width=0.9\textwidth]{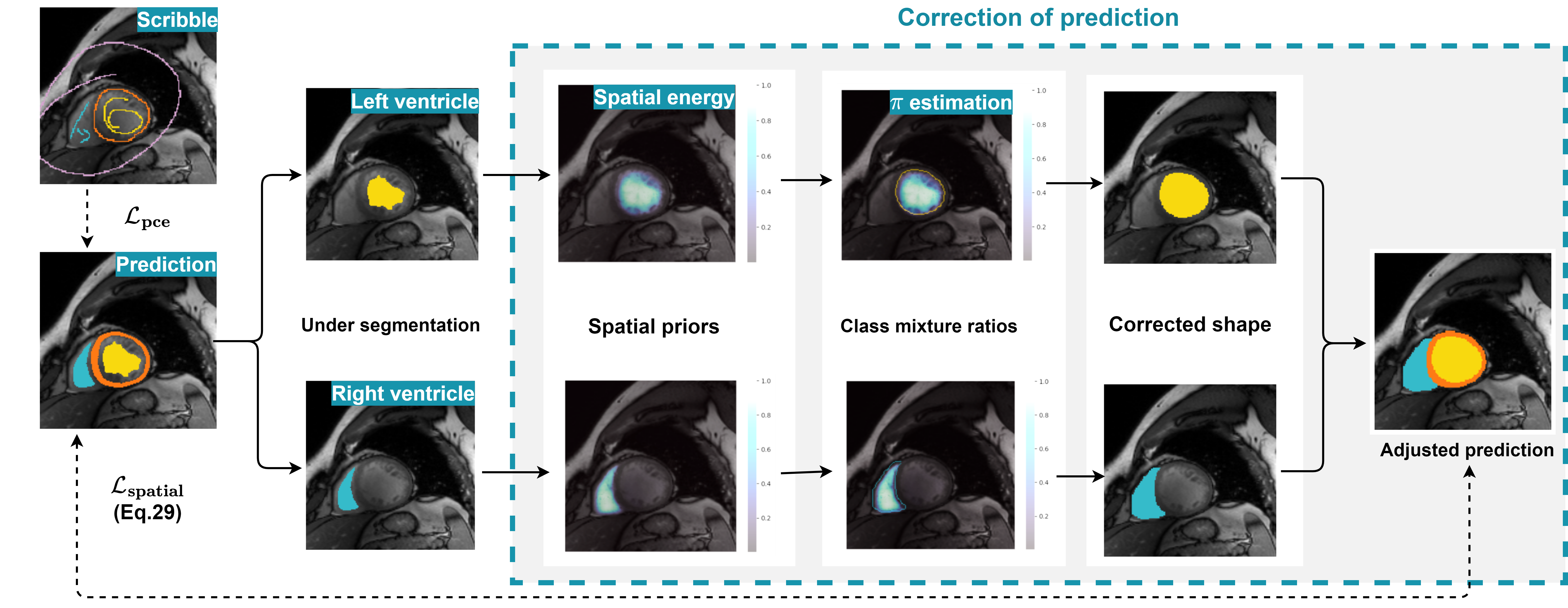}
    \caption{Illustration of the spatial prior loss ($\mathcal{L}_{\text{spatial}}$) for correction of prediction, via class mixture ratios ($\bm{\pi}$) and the spatial prior (with spatial energy). 
    }
    \label{fig:spatial}
\end{figure*}

To further exlploit the shape features and tackle the under-segmentation problem, as \zxhreffig{fig:roadmap} shows, we model class mixture ratios, the spatial prior, and the shape constraints to better capture shape information and regularize the network training.
As visualized in \zxhreffig{fig:spatial}, we compute the spatial energy to reflect the probabilities of pixels belonging to each class.
We seek to estimate the critical prior of label class proportions, referred to as $\bm{\pi}$, which guides the correction of erroneous network prediction.

\subsubsection{Problems statement}
\label{problems}
The segmentation network trained with scribbles tends to generate under segmentation results of the target structures. 
Considering that the annotated ratio of classes can be imbalanced, the scribble-supervised learning also brings challenges to the estimation of class mixture ratios $\bm{\pi}$.

\noindent\newline\textbf{Imbalanced classes}: Firstly, the number of pixels of different label class (denoted using $n_k$, where $N=\sum^m_{k=1} n_k$) in an image can be different. We denote this ratio of label class as  class mixture ratios, $\bm{\pi}=\{\pi_k\}^m_{k=1}$, and $n_k=N\times\pi_k$. 
Secondly, the ratios of scribble-annotated pixels to all the pixels of a label class, denoted as $\bm{a}=\{a_k\}^m_{k=1}$, are generally different.
In fully supervised segmentation, imbalanced classes from different class mixture ratios generally do not represent a challenge, as this ratio for each training image is known from the gold standard segmentation ($\pi_k=n_k/N$), where one can also view the annotation ratios of all labels are all equal to one, i.e., $a_{k}=1, k=1,\ldots,m$.
However, in scribble supervision we only have information about the scribble-annotated pixels of each class in a training image. 
Let us denote the numbers of annotated pixels of label classes as $\bm{n}_l=\{n_l^k\}^m_{k=1}$, where $n_l^k=n_k\times a_k$  and $n_k=N \times \pi_k$. 
Without the information of either mixture ratio $\pi_k$ or annotation ratio $a_k$, the model could not learn the coverage (number of pixels, $n_k$) for each class during training, thus easily leading to the phenomenon of under segmentation for certain classes.

\noindent\newline\textbf{Under segmentation:} 
As \zxhreffig{fig:underseg} shows, under segmentation refers to the results, where the size of segmented structure is generally smaller than ground truth, a phenomenon caused by the imbalanced annotated proportion and missed shape information. {Note that Over-segmentation is inherently unlikely to occur in scribble-supervised settings because sparse annotations provide only positive evidence for foreground. Since unlabeled pixels are not treated as foreground, it is hard to encourage the model to expand the predicted region beyond the foreground structures during training.}
To solve the problem, we propose to evaluate $\bm{\pi}$ and the spatial prior, which are crucial for the shape refinement.
The accurate estimation of $\bm{\pi}$ can correct the imbalanced label ratios, and enable model to adjust the size of segmentation result.
The computation of spatial prior is able to encode the feature similarity between pixels, and rectify the shape of target structures. 
We encode $\bm{\pi}$ and the spatial prior with a spatial prior loss, by ranking the spatial energy and select the top $\bm{\pi}$ ratio as the segmentation. 
To estimate $\bm{\pi}$, we start from the imbalanced class frequencies and adapt it from labeled pixels to unlabeled pixels.

Note that the problem of under segmentation can be even worse without the modeling of efficient scribbles.
In the case of manually annotated scribbles, the resulting annotations may be distributed in a non-random pattern due to fixed labeling habits, resulting in the biased label distribution across the whole dataset.
This problem could be alleviated by simulating randomly distributed labels through our proposed supervision augmentation.

\noindent\newline\textbf{Challenges of $\bm{\pi}$ estimation:} The evaluation of class mixture ratios is a critical bottleneck in semi-/ weak-/ non-supervised learning, and serves as the basis of classes identification~\citep{garg2021mixture} and variance reduction~\citep{wu2022minimizing,sakai2017semi}.
However, existing methods are mainly proposed for binary classification, and can not be adapted to multi-class scenario directly.
For segmentation task, the class mixture ratios are both imbalanced and interdependent, leading to the decrease in the performance of previous binary estimation approaches.
Despite the class imbalance problem, the scribble-supervised segmentation is also faced with the imbalance of annotated class ratios.
For example, the annotated ratio of the background tends to be much smaller than that of the foreground classes.
The imbalance of annotated ratio further enhances the difficulty of  $\bm{\pi}$ estimation.

\subsubsection{Estimation of class mixture ratios}
\label{Sec.estimation}
To mitigate under-segmentation in scribble-supervised learning, we estimate the class mixture ratios of the unlabeled pixels. Let $\bm{\pi}=[\pi_1,\ldots,\pi_m]$ denote the class prior over unlabeled pixels, where $\pi_k = p_u(c_k)$.

\noindent\newline\textbf{Objective:}
Given $n_u$ unlabeled pixels $\bm{x}=[x_1,\ldots,x_{n_u}]$ sampled from $p_u(x)$, we estimate $\bm{\pi}$ by maximizing the marginal likelihood of unlabeled observations:
\begin{linenomath*}\begin{equation*}
\mathcal{L}(\bm{\pi})=\prod_{i=1}^{n_u}p_u(x_i)
=\prod_{i=1}^{n_u}\left[\sum_{k=1}^m p_u(x_i\mid c_k)\,p_u(c_k)\right].
\label{eq_likelihood}
\end{equation*}\end{linenomath*}

We assume that labeled and unlabeled pixels share the same class-conditional appearance distribution (Assumption~\ref{ass:density-matching}), which is standard when scribbles are sampled without bias within each class.
\begin{assumption}[Within-class density matching between labeled and unlabeled pixels]
\label{ass:density-matching}
For each class $c_k$, we assume
\begin{equation}
p_\ell(x\mid c_k)=p_u(x\mid c_k)=p(x\mid c_k).
\end{equation}
\end{assumption}
This assumption holds when scribbles are sampled at random within each class. It may be violated under habitual or non-random scribbles (e.g. always tracing only the centre lines of structures or avoiding boundary regions). Our supervision augmentation (random mixup and occlusion) could mitigate such biases by randomizing the spatial distribution of labeled pixels.

\noindent\newline\textbf{EM formulation:}
We follow an EM procedure~\citep{latinne2001adjusting,mclachlan2007algorithm} to iteratively estimate $\bm{\pi}$. In the E-step, we compute the posterior responsibility of class $c_k$ for each unlabeled pixel $x_i$ using the current estimate $\bm{\pi}^{[t]}$. Let $\hat{p}_\ell(c_k\mid x_i)$ denote the network prediction (posterior) and $\hat{p}_\ell(c_k)$ denote the empirical class frequency on labeled pixels. Under Assumption~\ref{ass:density-matching}, the adapted posterior for unlabeled pixels is:
\begin{linenomath*}\begin{equation}
\hat{p}^{[t]}_u(c_k\mid x_i)
=
\frac{\pi^{[t]}_k\,\hat{p}_\ell(c_k\mid x_i)/\hat{p}_\ell(c_k)}
{\sum_{j=0}^{m}\pi^{[t]}_j\,\hat{p}_\ell(c_j\mid x_i)/\hat{p}_\ell(c_j)}.
\label{eq:posterior_adapt_final}
\end{equation}\end{linenomath*}

\noindent\newline\textbf{M-step:}
We then update the mixture ratio by averaging responsibilities over all unlabeled pixels:
\begin{linenomath*}\begin{equation}
\pi_k^{[t+1]}=\frac{1}{n_u}\sum_{i=1}^{n_u}\hat{p}^{[t]}_u(c_k\mid x_i).
\label{M_objective}
\end{equation}\end{linenomath*}

We initialize $\bm{\pi}^{[0]}$ with the class frequency on labeled pixels, i.e., $\hat{p}_\ell(c_k)=n_\ell^k/n_\ell$, and alternate Eq.~(\ref{eq:posterior_adapt_final}) and Eq.~(\ref{M_objective}) until convergence. Appendix~\ref{App:EM_mixture} provides the complete-data likelihood, the $Q$-function, and the derivation of Eq.~(\ref{eq:posterior_adapt_final}).

\noindent\newline\textbf{Discussion:} There are two conditions of the proposed algorithm. Firstly, we assume the within-class probabilities of labeled and unlabeled pixels be the same, which means the labeled pixels should be randomly sampled based on classes. 
Secondly, $\bm{\pi}$ is initiated with the class frequency of labeled pixels.
Since the annotated ratio of background is smaller than that of the foreground classes, the priori probabilities of foreground classes within unlabeled pixels tend to be over-estimated.
The first problem can be tackled by modeling the efficient scribbles, to achieve the random distribution of annotations. 
For the second problem, by randomly occluding the image and replacing the occluded area with background, we are able to increase the ratio of background and alleviate this problem to some extent. Furthermore, we propose to address it with the marginal probability maximization, which will be explained in Section~\ref{marginal}. Note that classical PU learning are primarily designed for binary settings and assume sample-wise independence. In our setting, the output space is multi-class, and the unlabeled set consists of densely correlated pixels within an image. Correspondingly, existing PU learning methods for classiﬁcation task cannot be directly used for multi-class segmentation. In this work, we adapt a PU-learning method nnPU~\citep{NIPS2017_7cce53cf} to the multi-class segmentation scenario and compare our approach with nnPU. Specifically, positive samples of each class are distinguished from unlabeled pixels. And then we maximize the marginal probability of negative samples.

\subsubsection{Computation of spatial energy}
Given the estimated class mixture ratios, we aim to identify the unlabeled pixels by determining the probability of pixels belonging to each class.
Instead of using model predictions directly, we further encode the spatial relationship to compensate the inaccurate results generated by segmentation network.
Inspired by \cite{Obukhov2019GatedCL}, we estimate the spatial energy of unlabeled pixels with energy term in a dense setting.

Firstly, we use Gaussian kernels $G_{ij}$ to measure the distance between pixels at position $i$ and $j$ as:
 
	\begin{linenomath*}\begin{equation}
	   G_{ij}  =\exp\left\{-\frac{(p_i-p_j)^2}{2\sigma_p^2}-\frac{(o_i-o_j)^2}{2\sigma_o^2}\right\},
       \label{3.3.3eq27}
	\end{equation}\end{linenomath*}
where $p_i$ represents the position of pixel $x_i$; $o_i$ denotes the intensity feature; $\sigma_p$ and $\sigma_o$ are the bandwidth parameters for position and intensity information, respectively. The shallow features like intensity and position are specific to the pixel and do not rely on the network prediction. Then, the energy term $\phi_{ij}$ leveraging prediction $\hat{y}$ is formulated as:
 
	\begin{linenomath*}\begin{equation*}
	    \phi_{ij}(\hat{y}) = G_{ij}\hat{y}_i\hat{y}_j,
	\end{equation*}\end{linenomath*}
which denotes the pairwise relationship between two pixels. This energy term connects every pixels with each other within one image. Based on $\phi_{i,j}$, we define the element of spatial energy $\Phi$ in a dense setting, \emph{i.e.},

	\begin{linenomath*}\begin{equation}
	    \Phi_i(\hat{y}) = \sum_{j\in\Omega_i} \phi_{ij}(\hat{y}),
	    \label{3.3eq11}
	\end{equation}\end{linenomath*} 
where $\Omega_i = \{\text{Pos}(i)-\text{Pos}(j)\leq r\}$, means the neighborhood window of radius $r$. Instead of taking the total energy as the regularization loss as~\citep{Obukhov2019GatedCL}, we consider $\Phi$ as the spatial energy to reflect the relative probability of pixels belonging to each class. 
\subsubsection{Spatial prior and shape regularization losses}
    \label{marginal}
\textbf{The spatial prior loss} is computed by ranking the spatial energy and selecting the top $\bm{\pi}$ proportion of pixels as the segmentation. 
Considering that adjusting multiple structures directly can be challenging, we instead separate each foreground class from the others, and then tackle the individual structure. 
Given that the mixture ratios of foreground classes tend to be over-estimated, we instead leverage the accurate negative pixels filtered by estimated mixture ratios, and maximize the marginal probability of these pixels belonging to other classes.

Firstly, by ranking the spatial energy and applying the mixture ratio of each class, we are able to distinguish negative pixels from unlabeled pixels. 
For foreground class $c_k$, we rank the unlabeled pixels according to the spatial energy $\Phi^{k}$ of class $c_k$ in Eq.~(\ref{3.3eq11}). 
Given the estimated mixture ratio $\pi_k$, we set pixels in the top  $\pi_k$ proportion to be positive samples $\Omega_k$.
Correspondingly, the remaining pixels are taken as negative pixels, denoted as $\bar{\Omega}_k$.
Taking over-estimated $\pi_k$ into account, we believe the set of negative pixels $\bar{\Omega}_k$ is more accurate than $\Omega_k$.
    
Secondly, we design the spatial prior loss ($\mathcal{L}_{\text{spatial}}$) based on maximal marginal probability of negative samples $\bar{\Omega}_k$ belonging to other classes.
For each class $c_k$, we take it as foreground and fuse other classes except $c_k$ into background. 
The fused class is denoted as $\bar{c}_k$.
For pixel $x_i$ in $\bar{\Omega}_k$, its marginal probability belonging to $\bar{c}_k$ equals the sum of probabilities of the fused classes, \emph{i.e.}, $\hat{p}(\bar{c}_k|x_i, x_i \in \bar{\Omega}_k) = \sum_{k'=1}^m [\mathbf{1}_{[k'\neq k]}\hat{p}(c_k|x_i)]$.
To maximize the marginal probability of negative pixel $x_i$ belonging to $\bar{c}_k$, we formulate the spatial prior loss as:
    
    \begin{linenomath*}\begin{equation}
        \mathcal{L}_{\text{spatial}} =-\sum_{k=1}^m \sum_{x_i\in \bar{\Omega}_k}\log(\hat{p}\left(\bar{c}_k|x_i)\right).
        \label{eq19}
    \end{equation}\end{linenomath*}

Our construction involves selecting a fixed-coverage pixel subset via sorting, which is different of region-selection mechanisms used in perturbation-based explanation methods~\citep{Fong_iccv17, Fong_iccv19}. However, unlike~\cite{Fong_iccv17, Fong_iccv19} we do not optimize a perturbation mask nor aim at post-hoc interpretability. Instead, we rank a class-wise spatial energy map and use the estimated mixture ratio $\pi_k$ to define class-specific sets, then regularize predictions through a probabilistic constraint to reduce under-segmentation under sparse scribble supervision.

\noindent\textbf{The shape regularization loss}. Due to insufficient supervision of scribbles, the network often produces multiple disconnected regions. To mitigate this problem, we propose $\mathcal{L}_{\text{shape}}$ to regularize the inter-connectivity within the structure. Different with the spatial prior loss, which constrains the size of the structure and the correlation between pixels.
We adopt $\mathcal{L}_{\text{shape}}$ to further reduce noise and smooth boundary.
It requires the model prediction to be consistent with its maximum connected area, and minimizes their cross entropy loss, \emph{i.e.},
    
    \begin{linenomath*}\begin{equation}
    \small
        \mathcal{L}_{\text{shape}} = -\sum_{k \in \varPsi}F(\hat{Y}_k)\log(\hat{Y}_k),
    \label{4.2eq1}
    \end{equation}\end{linenomath*}
where $\varPsi$ is the set of label classes with inter-connected structures; $F(\cdot)$ denotes the morphological function, and outputs the largest inter-connected area of input label. For targets that naturally contain multiple non-connected regions (e.g., multifocal lesions), $\mathcal{L}_{\text{shape}}$ should be disabled for the corresponding class, because the morphological operation $F(\cdot)$ would remove valid target regions.
 

\subsection{ZScribbleNet}
ZScribbleSeg is achieved via a deep neural network referred to as ZScribbleNet. Our goal is to develop a learning framework for scribble supervision by efficiently modeling scribbles and correcting network predictions via prior regularization. The proposed method does not depend on particular network architectures and can be directly applied to any segmentation backbone.
To demonstrate the effectiveness of our method while decoupling it from architectural improvements, we adopted the UNet \citep{baumgartner2017exploration} as the backbone for all experiments.

As \zxhreffig{fig:overview} shows,
two images are mixed together to perform the supervision augmentation.
Then, our ZScribbleNet takes the mixed images and unmixed images as the input, and output their segmentation results.

For model training, images and their scribble annotations are sampled to estimate the training objective ($\mathcal{L}$), which is formulated as:

\begin{linenomath*}\begin{equation}
    \mathcal{L} = \mathcal{L}_{\text{pce}}+\underbrace{\lambda_1\mathcal{L}_{\text{global}}+\lambda_2\mathcal{L}_{\text{spatial}} +\lambda_3\mathcal{L}_{\text{shape}}}_{\text{unsup}},
\label{eq:final}
\end{equation}\end{linenomath*}

where $\mathcal{L}_{\text{pce}}$ is the partial cross entropy loss calculated for annotated pixels in unmixed image and mixed image. The global consistency loss $\mathcal{L}_{\text{global}}$ in Eq.(\ref{4.1eq4}) requires the mix equivalence for the supervision augmentation. The spatial prior loss $\mathcal{L}_{\text{spatial}}$ in Eq.(\ref{eq19}) encodes the $\bm{\pi}$ prior and spatial prior; shape regularization loss$\mathcal{L}_{\text{shape}}$ in Eq.(\ref{4.2eq1}) leverages shape constraint. $\lambda_{1},\lambda_2,\lambda_3$ are hyper-parameters to leverage the relative importance of different loss components. \zxhreftb{tab:loss-hyperparams} summarized the hyper-parameters used in our method.

\begin{table}[t]
\caption{Summary of hyper-parameters used in the proposed training losses.}
\label{tab:loss-hyperparams}
\centering
\setlength{\tabcolsep}{6pt}
{
\begin{tabular}{ll}
\toprule
Notations & Description \\
\midrule
$\lambda_1$ & Weight of the global consistency loss $\mathcal{L}_{\text{global}}$. \\
$\lambda_2$ & Weight of the spatial prior loss $\mathcal{L}_{\text{spatial}}$. \\
$\lambda_3$ & Weight of the shape regularization loss$\mathcal{L}_{\text{shape}}$. \\
$s_b$ & Side length of the occlusion mask in $\mathbf{1}_b$. \\
$\sigma_p$ & Position bandwidth of Gaussian kernel in Eq.~\ref{3.3.3eq27}. \\
$\sigma_o$ & Intensity bandwidth of Gaussian kernel in Eq.~\ref{3.3.3eq27}. \\
$r$ & Neighborhood radius used in Eq.~\ref{3.3eq11}. \\
$E_{\text{warm}}$ & Warm-up epochs before enabling $\mathcal{L}_{\text{spatial}}$. \\
\bottomrule
\end{tabular}
}
\end{table}

At the beginning of training, the prediction results of the model could be inaccurate, which would introduce noise to the estimation of class mixture ratios $\bm{\pi}$ and spatial energy. Therefore, we warmly started training the networks with partial cross entropy loss $\mathcal{L}_{\text{pce}}$, global consistency loss $\mathcal{L}_{\text{global}}$, and shape regularization loss $\mathcal{L}_{\text{shape}}$ for 100 epochs, and then invoked the spatial loss $\mathcal{L}_{\text{spatial}}$.

In the testing phase, the trained network predicted the segmentation results of input image directly.

\section{Experiments and Results}
We first introduce the materials used in our study, including four segmentation tasks, datasets, and evaluation protocols (Section~\ref{sec-exp-materials}).
Then, we investigate a variety of scribble forms and analyze the principles of efficient scribbles (Section~\ref{sec-exp-scribble}).
Next, we perform an ablation study to assess the contribution of each component in the proposed ZScribbleSeg framework (Section~\ref{sec-exp-ablation}).
Finally, we compare ZScribbleSeg with state-of-the-art methods across diverse segmentation tasks using four open datasets (Section~\ref{sec-exp-sota}).

\subsection{Materials}\label{sec-exp-materials}

\subsubsection{Tasks and datasets}
Our validation covered four representative segmentation tasks: (1) cardiac ventricular segmentation of regular structures from anatomical MRI (ACDC) and abdomen organ structure segmentation from CT (BTCV), (2) regular structure segmentation from pathology-enhanced imaging with limited training samples (MSCMRseg), (3) irregular myocardial pathology segmentation (MyoPS) and brain tumor segmentation (Decathlon-BrainTumor)from multi-modality MRI (MyoPS), and (4) 3D prostate segmentation from multi-modality MRI (Decathlon).

\textbf{ACDC} dataset was from 
the MICCAI’17 Automatic Cardiac Diagnosis Challenge~\cite{8360453}. 
This dataset consists of short-axis cardiac images using anatomical MRI sequence (BSSFP) from 100 patients, with gold standard segmentation of cardiac ventricular structures, including left ventricle blood cavity (LV), left ventricle myocardium (MYO), and right ventricle blood cavity (RV).
For experiments, we randomly divided the 100 subjects into a training set of 70 subjects, a validation set of 15  subjects (particularly for ablation study), and a test set of 15  subjects.

\textbf{MSCMRseg} was from the MICCAI’19 Multi-sequence Cardiac MR Segmentation Challenge~\citep{8458220,Zhuang2016MultivariateMM}, consisting of images from 45 patients with cardiomyopathy and the gold standard segmentation of LV, MYO and RV. 
We employed the 45 images of late gadolinium enhanced (LGE) MRI to evaluate the segmentation of ventricle structures.
Following \cite{yue2019cardiac}, we divided the 45 images into three sets of 25 (training), 5 (validation), and 15 (test) images for all experiments.
Note that this structure segmentation is more challenging than that on ACDC due to its smaller training set and pathology enhanced images.

\textbf{MyoPS} dataset was from MICCAI'20 Myocardial pathology segmentation Challenge~\citep{li2022myops}, consisting of paired images of BSSFP, LGE and T2 cardiac MRI from 45 patients. 
The task was to segment the myocardial pathologies, including scar and edema, which do not have regular shape or structure thus their segmentation represents a different task to the regular structure segmentation. 
Following the benchmark study~\citep{li2022myops}, we split the data into 20 pairs of training set, 5 pairs of validation set and 20 pairs of test set.

\textbf{Decathlon-Prostate} dataset was released by the Decathlon challenge~\citep{antonelli2022medical}. The organizers released 32 paired MRI 3D volumes of two series, including transverse T2-weighted and the apparent diffusion coefficient (ADC). The ground truth segmentation of central gland and peripheral zone were also provided along with the images.
We excluded two cases with missing labels of peripheral zones and split the remaining 30 paired volumes into three sets of 18(training), 6(validation) and 6(test). To avoid the deviation caused by random dataset division, we conducted 5-fold cross-validation and reported the average performance on test set.

\textbf{Decathlon-BrainTumor} dataset consists of 750 multiparametric-magnetic resonance images from patients diagnosed with either glioblastoma or lower-grade glioma~\citep{Antodechathlon}. The Decathlon-BrainTumor dataset contains the same cases as the 2016 and 2017 Brain Tumor Segmentation (BraTS) challenges. They have a uniform resolution with a size of $240\times240\times155$. We aimed to segment two foreground classes: the tumor core and the peritumoral edema. The dataset was randomly split into 80\%, 10\% and 10\% for training, validation and testing, respectively.

\textbf{BTCV} dataset comprises 47 abdominal CT acquired at the Vanderbilt University Medical Center from metastatic liver cancer patients or post-operative ventral hernia patients~\cite{landman2015miccai}. We aimed to segment three organs: the liver, spleen and stomach. Same as Decathlon-BrainTumor dataset, the BTCV dataset was randomly split into 80\%, 10\% and 10\% for training, validation and testing, respectively.

\subsubsection{Evaluation metrics} 
For each task, we reported the Dice score and Hausdorff Distance (HD) on each foreground class separately. 

\begin{table*}[!thb] 
\caption{Efficiency analysis of scribble forms for regular structure segmentation of cardiac ventricles (ACDC dataset) and irregular segmentation of myocardial pathology (MyoPS dataset). Here,
$N_\text{scribble}$ and $N_\text{pix}$ respectively denote the number of manual draws to generate scribble annotations and number of annotated pixels, which indicate annotation efforts; 
$\bm{d}=\{d_k\}_{k=1}^m$ is the number of manual draws (scribbles) and $\bm{n}_l=\{n^k_l\}_{k=1}^m$ is the given threshold of annotation efforts (number of labeled pixels), where $d_k<<n^k_l$ for each class k. 
Segmentation results are evaluated on test set and reported in Dice scores.
}\label{tab1}
\centering
\resizebox{\textwidth}{!}{
\begin{tabular}{ccc cccc ccc}
\hline  \multirow{2}{*}[-4pt]{Methods}&\multirow{2}{*}[-4pt]{$N_\text{scribble}$}&\multirow{2}{*}[-4pt]{$N_\text{pix}$}&\multicolumn{4}{c|}{Structural segmentation}&\multicolumn{3}{c}{Irregular segmentation}\\
\cmidrule(lr){4-7}\cmidrule(lr){8-10}
&&&LV & MYO & RV & \multicolumn{1}{c|}{Avg} & Scar & Edema &\multicolumn{1}{c}{Avg} \\
\hline

\multicolumn{1}{c|}{Skeleton}&$\bm{d}$&\multicolumn{1}{c|}{$\bm{n}_l$}&.805$\pm$.145&.737$\pm$.095&.769$\pm$.128&\multicolumn{1}{c|}{.770$\pm$.126}& 
.504$\pm$.213&.057$\pm$.022&\multicolumn{1}{c}{.281$\pm$.271} \\

\multicolumn{1}{c|}{Random walk}&$\bm{d}$&\multicolumn{1}{c|}{$\bm{n}_l$}&.798$\pm$.173&.698$\pm$.153&.753$\pm$.157&\multicolumn{1}{c|}{.744$\pm$.165}&
.516$\pm$.284&.529$\pm$.123&\multicolumn{1}{c}{.522$\pm$.184}\\

\multicolumn{1}{c|}{DirRandomWork}&$\bm{d}$&\multicolumn{1}{c|}{$\bm{n}_l$}&\underline{.844$\pm$.143}&\underline{.755$\pm$.102}&\underline{.798$\pm$.173}&\multicolumn{1}{c|}{\underline{.799$\pm$.146}}&
\underline{.539$\pm$.217}&\underline{.637$\pm$.108}&\multicolumn{1}{c}{\underline{.588$\pm$.176}}\\

\multicolumn{1}{c|}{Points}&$\bm{n}_l$&\multicolumn{1}{c|}{$\bm{n}_l$}&\textbf{.876$\pm$.134}&\textbf{.801$\pm$.089}&\textbf{.858$\pm$.081}&\multicolumn{1}{c|}{\textbf{.845$\pm$.107}}&
\textbf{.551$\pm$.246}&\textbf{.638$\pm$.115}&\multicolumn{1}{c}{\textbf{.595$\pm$.194}}\\
  
\hline
\end{tabular}}
\end{table*} 

\begin{figure*}[t]
\centering
\subfigure[]{
\includegraphics[width=0.24\textwidth]{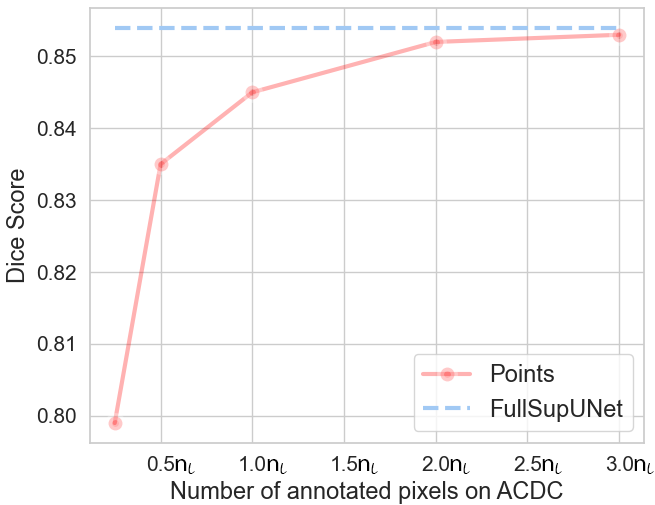}}
\subfigure[]{
\includegraphics[width=0.24\textwidth]{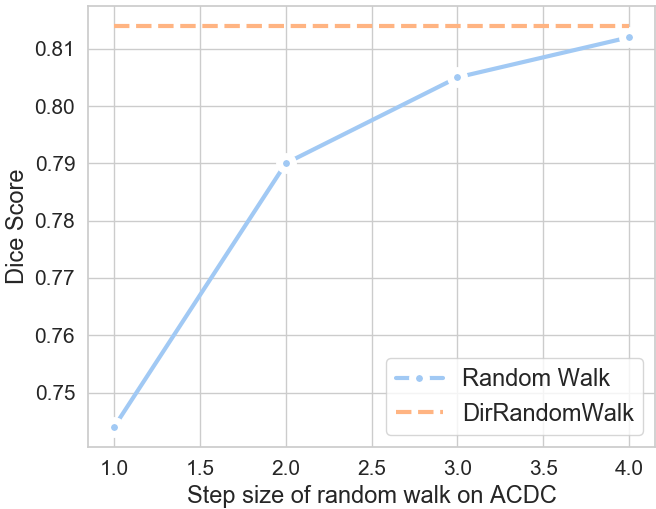}}
\subfigure[]{
\includegraphics[width=0.23\textwidth]{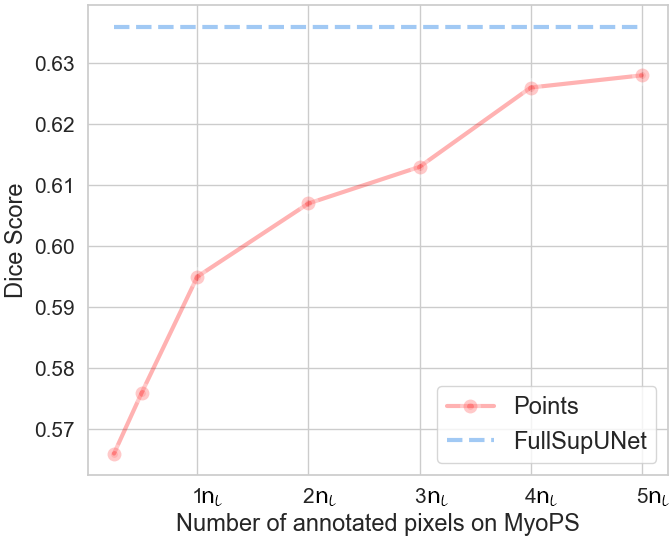}}
\subfigure[]{
\includegraphics[width=0.24\textwidth]{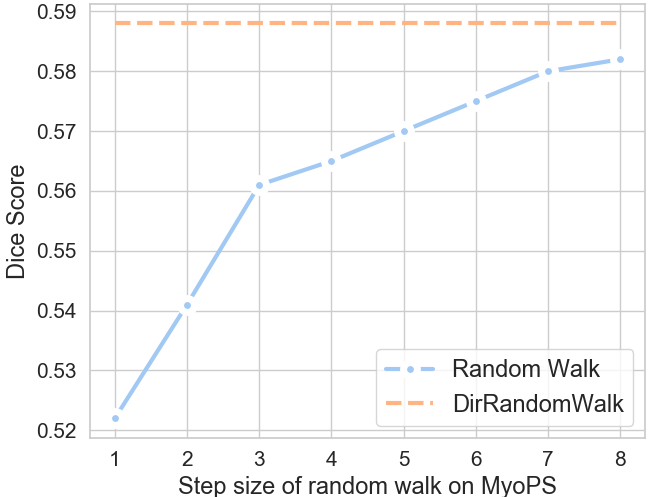}}
\caption{Performance of segmentation networks trained using partial cross entropy computed from Points scribble form with different number of pixels $N_{pix}$, with comparisons to fully supervised models (FullSupUNet): (a) and (c) visualize Dice scores with respect to different $N_{pix}$ on ACDC and MyoPS, respectively. 
The performance of models trained by the Random walk form, with increasing step length $l$, compared with models trained by DirRandWalk:  (b) and (d) show the Dice scores of segmentation on ACDC and MyoPS, respectively, given $N_{pix}=\bm{n}_l$.}
\label{fig:exp:scribbleform}\end{figure*}

\subsubsection{Pre-processing and implementation}

The two dimensional slices from ACDC and MSCMR datasets were of different resolutions. Hence, we first re-sampled all images into a fixed resolution of $1.37\times1.37$~mm and then extracted the central patch of size $212\times212$ for experiments.
For MyoPS, we took the paired slices of BSSFP, LGE, and T2 CMR and cropped their center patches of size $192\times192$ for experiments.
For Prostate-Decathlon, we re-sampled all volumes to the resolution of $1.1\times1.1\times1.1$~mm. We then cropped $64\times64\times64$ center patches for each volume and take a paired patches of T2 and ADC sequences as an input. 
We normalized the intensity of these medical images to be zero mean and unit variance. For BTCV dataset, we resampled all images into a fixed resolution of $1.2\times1.2$~mm and resized the images to $256\times256$. For Decathlon-BrainTumor dataset, we used the original image resolution for our experiments.

For random occlusion, a square area of $32\times 32$ was randomly occluded for each image.
For the estimation of spatial energy, We adopted Gaussian kernels with intensity bandwidth $\sigma_{o} = 0.1$, position bandwidth $\sigma_p = 6$, and kernel radius $r=5$.
The hyper-parameters $\lambda_1$, $\lambda_2$, $\lambda_3$ in Eq.~(\ref{eq:final}) were empirically set to be $0.05$, $1$, and $1$, respectively.

All models were trained with a batch size of 4, learning rate of 1e$^{-4}$, and augmentation of flipping and random rotation.    
We implemented our models with Pytorch. All models were trained on one NVIDIA 3090Ti 24GB GPU for 1000 epochs.

\subsection{Efficiency of scribble forms}
\label{sec-exp-scribble}


In this study, we first compared four scribble forms to illustrate the efficacy of randomly annotated scribbles for supervision.
Denoting the number of annotated pixels using a manual and skeleton-wise scribble form as $n$, we generated other scribble forms with the same annotated ratios for a fair comparison.
Then, we studied the performance of segmentation with respect to the number of pixels annotated using a random and wide range scribble form, by setting the number of annotated pixels to different times of $n$.
Finally, we further explored variants of random walk annotations to show the importance of wide range in the random distribution of scribbles. 

We applied two segmentation tasks, \ie regular structure segmentation of the cardiac ventricles on ACDC dataset and irregular segmentation of myocardial pathologies using MyoPS dataset. 
To compare the supervision of scribble forms directly, we trained all models with partial cross entropy (PCE) loss calculated for annotated pixels from scribbles.
All experiment results were reported on the test set.

\begin{table*}[!h]
\caption{Results in Dice scores and Hausdorff Distance (HD) of the ablation study using ACDC dataset, where the models were evaluated on the validation set. Note that model~\#8 is  ZScribbleSeg.
\textbf{Bold} denotes the best result, and \underline{underline} indicates the best but one in each category.}\label{tab3}
\centering
\resizebox{\textwidth}{!}{
\begin{tabular}{c|ccccccc|cccc}
\hline 
\multirow{2}{*}{Dice}&\multirow{2}{*}{$\mathcal{L}_{\text{pce}}$} &\multicolumn{2}{c}{\textcolor{black}{Efficiency}}&\multirow{2}{*}{$\mathcal{L}_{\text{global}}$}&\multirow{2}{*}{$\mathcal{L}_{\text{shape}}$}&\multicolumn{2}{c|}{$\textcolor{black}{\mathcal{L}_{\text{spatial}}}$}& \multirow{2}{*}{LV} &\multirow{2}{*}{MYO} & \multirow{2}{*}{RV} &\multirow{2}{*}{Avg}\\
\cmidrule(lr){3-4}\cmidrule(lr){7-8}
&&\textcolor{black}{Mix} & \textcolor{black}{Occlusion}&&&\textcolor{black}{$\bm{\pi}$}&\textcolor{black}{\text{spatial}}&&&&\\
\hline
model~{\#1}&$\checkmark$&$\times$&$\times$&$\times$&$\times$&$\times$&\multicolumn{1}{c|}{$\times$}&.863$\pm$.089&.804$\pm$.063&.774$\pm$.150&.813$\pm$.112\\
\textcolor{black}{model~{\#2}}&\textcolor{black}{$\checkmark$}&\textcolor{black}{$\checkmark$}&\textcolor{black}{$\times$}&\textcolor{black}{$\times$}&\textcolor{black}{$\times$}&\textcolor{black}{$\times$}&\multicolumn{1}{c|}{\textcolor{black}{$\times$}}&\textcolor{black}{.871$\pm$.098}&\textcolor{black}{.811$\pm$.096}&\textcolor{black}{.771$\pm$.197}&\textcolor{black}{.818$\pm$.143}\\
\textcolor{black}{model~{\#3}}&\textcolor{black}{$\checkmark$}&\textcolor{black}{$\checkmark$}&\textcolor{black}{$\checkmark$}&\textcolor{black}{$\times$}&\textcolor{black}{$\times$}&\textcolor{black}{$\times$}&\multicolumn{1}{c|}{\textcolor{black}{$\times$}}&\textcolor{black}{.870$\pm$.100}&\textcolor{black}{.833$\pm$.063}&\textcolor{black}{.843$\pm$.076}&\textcolor{black}{.848$\pm$.082}\\
model~{\#4}&$\checkmark$&$\checkmark$&$\checkmark$&$\checkmark$&$\times$&$\times$&\multicolumn{1}{c|}{$\times$}&.920$\pm$.064&.868$\pm$.051&.886$\pm$.051&.891$\pm$.059\\
model~{\#5}&$\checkmark$&$\times$&$\times$&$\times$&$\checkmark$&$\times$&\multicolumn{1}{c|}{$\times$}&.915$\pm$.068&.871$\pm$.056&.871$\pm$.058&.886$\pm$.064\\
\textcolor{black}{model~{\#6}}&\textcolor{black}{$\checkmark$}&\textcolor{black}{$\times$}&\textcolor{black}{$\times$}&\textcolor{black}{$\times$}&\textcolor{black}{$\times$}&\textcolor{black}{$\checkmark$}&\multicolumn{1}{c|}{\textcolor{black}{$\times$}}&\textcolor{black}{.919$\pm$.045}&\textcolor{black}{.855$\pm$.050}&\textcolor{black}{.869$\pm$.055}&\textcolor{black}{.881$\pm$.057}\\
\textcolor{black}{model~{\#7}}&\textcolor{black}{$\checkmark$}&\textcolor{black}{$\times$}&\textcolor{black}{$\times$}&\textcolor{black}{$\times$}&\textcolor{black}{$\times$}&\textcolor{black}{$\checkmark$}&\multicolumn{1}{c|}{\textcolor{black}{$\checkmark$}}&\textcolor{black}{\underline{.923$\pm$.078}}&\textcolor{black}{\underline{.869$\pm$.051}}&\textcolor{black}{\underline{.889$\pm$.056}}&\textcolor{black}{\underline{.894$\pm$.066}}\\
model~{\#8}&$\checkmark$&$\checkmark$&$\checkmark$&$\checkmark$&$\checkmark$&$\checkmark$&\multicolumn{1}{c|}{$\checkmark$}&\textbf{.929$\pm$.057}&\textbf{.876$\pm$.051}&\textbf{.892$\pm$.049}&\textbf{.899$\pm$.056}\\
\hline
HD(mm)&$\mathcal{L}_{\text{pce}}$ &\textcolor{black}{Mix}& \textcolor{black}{Occlusion}&$\mathcal{L}_{\text{global}}$&$\mathcal{L}_{\text{shape}}$&\textcolor{black}{$\bm{\pi}$}&\textcolor{black}{$\mathcal{L}_{\text{spatial}}$}&LV & MYO & RV &Avg\\
\hline
model~{\#1}&$\checkmark$&$\times$&$\times$&$\times$&$\times$&$\times$&\multicolumn{1}{c|}{$\times$}&81.86$\pm$40.40&65.97$\pm$33.62&60.91$\pm$44.62&69.58$\pm$40.37\\
\textcolor{black}{model~{\#2}}&\textcolor{black}{$\checkmark$}&\textcolor{black}{$\checkmark$}&\textcolor{black}{$\times$}&\textcolor{black}{$\times$}&\textcolor{black}{$\times$}&\textcolor{black}{$\times$}&\multicolumn{1}{c|}{\textcolor{black}{$\times$}}&\textcolor{black}{88.88$\pm$12.42}&\textcolor{black}{37.18$\pm$34.04}&\textcolor{black}{95.01$\pm$23.32}&\textcolor{black}{73.69$\pm$35.85}\\
\textcolor{black}{model~{\#3}}&\textcolor{black}{$\checkmark$}&\textcolor{black}{$\checkmark$}&\textcolor{black}{$\checkmark$}&\textcolor{black}{$\times$}&\textcolor{black}{$\times$}&\textcolor{black}{$\times$}&\multicolumn{1}{c|}{\textcolor{black}{$\times$}}&\textcolor{black}{119.78$\pm$19.14}&\textcolor{black}{23.90$\pm$17.32}&\textcolor{black}{52.38$\pm$23.40}&\textcolor{black}{65.35$\pm$45.06}\\
model~{\#4}&$\checkmark$&$\checkmark$&$\checkmark$&$\checkmark$&$\times$&$\times$&\multicolumn{1}{c|}{$\times$}&12.12$\pm$18.26&29.41$\pm$24.56&16.97$\pm$15.62&19.50$\pm$20.94\\
model~{\#5}&$\checkmark$&$\times$&$\times$&$\times$&$\checkmark$&$\times$&\multicolumn{1}{c|}{$\times$}&\textbf{4.45$\pm$5.39}&\underline{15.24$\pm$23.90}&25.78$\pm$22.44&\underline{15.16$\pm$20.89}\\
\textcolor{black}{model~{\#6}}&\textcolor{black}{$\checkmark$}&\textcolor{black}{$\times$}&\textcolor{black}{$\times$}&\textcolor{black}{$\times$}&\textcolor{black}{$\times$}&\textcolor{black}{$\checkmark$}&\multicolumn{1}{c|}{\textcolor{black}{$\times$}}&\textcolor{black}{39.95$\pm$37.55}&\textcolor{black}{50.32$\pm$36.14}&\textcolor{black}{42.49$\pm$31.10}&\textcolor{black}{44.25$\pm$34.93}\\
\textcolor{black}{model~{\#7}}&\textcolor{black}{$\checkmark$}&\textcolor{black}{$\times$}&\textcolor{black}{$\times$}&\textcolor{black}{$\times$}&\textcolor{black}{$\times$}&\textcolor{black}{$\checkmark$}&\multicolumn{1}{c|}{\textcolor{black}{$\checkmark$}}&\textcolor{black}{28.95$\pm$36.57}&\textcolor{black}{44.77$\pm$34.69}&\textcolor{black}{\textbf{7.51$\pm$5.34}}&\textcolor{black}{27.08$\pm$32.76}\\
model~{\#8}&$\checkmark$&$\checkmark$&$\checkmark$&$\checkmark$&$\checkmark$&$\checkmark$&\multicolumn{1}{c|}{$\checkmark$}&\underline{6.09$\pm$8.53}&\textbf{11.14$\pm$14.53}&\underline{8.86$\pm$5.88}&\textbf{8.70$\pm$10.40}\\
\hline
\end{tabular}}
\end{table*} 

\subsubsection{Scribble forms}  \label{sec:exp-scribbleform}
Annotation effort can be assessed in two ways. One is the number of manual draws used to create scribbles ($N_\text{scribble}$), and the other is the number of annotated pixels ($N_\text{pix}$).
Given the certain amount of efforts, we designed four forms following different generation procedures, including (1) Skeleton, (2) Random walk, (3) Directed random walk (DirRandomWalk), (4) Points. 
The details of scribble forms are described bellow.

\textbf{Skeleton} indicates the widely adopted scribble form by a annotator, who approximately outlines the shape of each label class within the segmentation mask. 
For a segmentation task with $m$ label classes, one needs $\bm{d}$ manual draws (scribbles) for a training image,
where $\bm{d}=\{d_k\}_{k=1}^m$ and $d_k\geq 1$ denotes the number of manual draws for label class $k$.
For ACDC dataset, we adopted the manual annotated skeleton scribble released by~\cite{9389796}.

\textbf{Random walk} starts from a random point within the segmentation mask.
Then, the annotation moves along a random direction of image lattice within the segmentation mask with a given step length ($l$ by default set to 1).
We repeated such moves until the ratio or number of annotated pixels reached a threshold ($\bm{n}_l$).

\textbf{Directed random walk} (DirRandomWalk) is a variant of the random walk with momentum. Scribbles generated by the standard random walk often cluster within a local region of radius $\sqrt{r}$ after $r$ steps. To promote broader coverage without manually specifying the step length ($l$), the directed random walk encourages movement in the same direction as the previous step. When the next point falls outside the segmentation mask, the direction is adjusted to the nearest feasible angle.

\textbf{Points} represents an idealized scribble form, where annotated pixels are randomly sampled within the segmentation mask. However, it is difficult to generate such scribble annotation in practice, due to the huge number of manual draws which equals the number of annotated pixels, that is,  $N_\text{scribble}=N_\text{pix}$. Therefore, we considered this form as the upper bound of scribble supervision under the same ratio of annotated pixels.

\subsubsection{Results} 
Given the same amount of annotated pixels, we show the effect of different scribble forms on regular structures (ACDC) and irregular objects (MyoPS).
As \zxhreftb{tab1} illustrates, when the four scribble forms had the same number of annotated pixels $N_\text{pix}$, Points achieved the best Dice scores on both of the structural segmentation and irregular segmentation tasks.This can be attributed to the effects of randomness and wide range distribution of scribbles. 
However, when the annotation effort was constrained by the number of manual draws, DirRandomWalk became more favorable. In this setting, the Points form was impractical despite its superior accuracy.
Furthermore, Skeleton was shown to be the least efficient form. The segmentation neural network trained on such annotation performed poorly on the irregular object segmentation task.
This was due to the fact that when the target was difficult to outline, Skeleton form could fail to portray the entire segmentation. Such incomplete annotations result in poor performance or even training failure.
By contrast, randomly distributed scribble forms, such as Random walk and DirRandomWalk, demonstrated clear superiority. On irregular object segmentation, they improved the average Dice score over Skeleton by $24.1\%$ and $30.7\%$, respectively.
\begin{table*}[!t]
\caption{Results and comparisons of regular structure segmentation on ACDC dataset.
}\label{tab5}
\centering
\resizebox{\textwidth}{!}{
\begin{tabular}{ccccccccc}
\hline
\multirow{2}{*}[-4pt]{Methods}&\multicolumn{4}{c|}{Dice}&\multicolumn{4}{c}{HD (mm)}\\
\cmidrule(lr){2-5}\cmidrule(lr){6-9}
&LV & MYO & RV &\multicolumn{1}{c|}{Avg} & LV & MYO & RV & Avg \\
\hline
\multicolumn{1}{c|}{PCE}&.805$\pm$.145&.737$\pm$.095&.769$\pm$.128&\multicolumn{1}{c|}{.770$\pm$.126}&62.55$\pm$36.04&68.30$\pm$27.77&59.62$\pm$42.62&63.40$\pm$35.76\\
\multicolumn{1}{c|}
{\textcolor{black}{WSL4~\citep{luo2022scribble}}}&\textcolor{black}{.890$\pm$.098}&\textcolor{black}{.818$\pm$.057}&\textcolor{black}{.863$\pm$.068}&\multicolumn{1}{c|}{\textcolor{black}{.857$\pm$.074}}&\textcolor{black}{39.75$\pm$47.27}&\textcolor{black}{68.80$\pm$49.66}&\textcolor{black}{31.61$\pm$29.93}&\textcolor{black}{46.72$\pm$42.29}\\
\multicolumn{1}{c|}{GatedCRF~\citep{Obukhov2019GatedCL}}&.846$\pm$.157&.744$\pm$.108&.822$\pm$.111&\multicolumn{1}{c|}{.804$\pm$.135}&37.38$\pm$46.37&22.30$\pm$15.72&20.88$\pm$11.85&26.85$\pm$30.03\\
\multicolumn{1}{c|}{MAAG~\citep{9389796}}&.879&.817&.752&\multicolumn{1}{c|}{.816}&25.23&26.83&22.73&24.93\\
\multicolumn{1}{c|}{CVIR~\citep{garg2021mixture}}&.866$\pm$.127&.797$\pm$.102&.737$\pm$.130&\multicolumn{1}{c|}{.800$\pm$.130}&47.51$\pm$50.82&10.70$\pm$8.39&14.39$\pm$9.00&.24.20$\pm$34.17\\	\multicolumn{1}{c|}{nnPU~\citep{NIPS2017_7cce53cf}}&.862$\pm$.134&.792$\pm$.124&.829$\pm$.102&\multicolumn{1}{c|}{.828$\pm$.123}&67.28$\pm$48.60&18.60$\pm$17.93&14.64$\pm$8.39&33.51$\pm$38.43\\		\multicolumn{1}{c|}{CycleMix~\citep{zhang2022cyclemix}}&.876$\pm$.096&.794$\pm$.083&.829$\pm$.099&\multicolumn{1}{c|}{.833$\pm$.098}&16.60$\pm$19.90&18.04$\pm$17.78&19.09$\pm$21.44&17.91$\pm$19.57\\
\multicolumn{1}{c|}{ShapePU~\citep{zhang2022shapepu}}&.885$\pm$.103&.806$\pm$.096&.851$\pm$.089&\multicolumn{1}{c|}{.848$\pm$.100}&20.17$\pm$22.40&41.81$\pm$33.40&20.06$\pm$26.43&27.35$\pm$29.33\\
\multicolumn{1}{c|}{TIP25~\citep{chentip25}}&
.894$\pm$.094&
.822$\pm$.059&
.873$\pm$.066&
\multicolumn{1}{c|}{\underline{.863$\pm$.061}}&
\underline{7.26$\pm$4.11}&
\underline{7.08$\pm$9.54}&
\underline{8.74$\pm$10.11}&
\underline{7.69$\pm$7.01}\\
\multicolumn{1}{c|}{HELPNet~\citep{zhangmedia25}}&
\textbf{.931$\pm$.043}&
\textbf{.859$\pm$.057}&
\textbf{.931$\pm$.041}&
\multicolumn{1}{c|}{\textbf{.901$\pm$.034}}&
\textbf{5.80$\pm$3.29}&
\textbf{2.78$\pm$2.16}&
\textbf{3.92$\pm$6.74}&
\textbf{3.83$\pm$3.48}\\
\multicolumn{1}{c|}{ScribFormer~\citep{li2024scribformer}}&
.880$\pm$.109&
\underline{.794$\pm$.078}&
\underline{.894$\pm$.063}&
\multicolumn{1}{c|}{.856$\pm$.072}&
14.96$\pm$18.25&
17.38$\pm$19.99&
13.48$\pm$8.03&
15.27$\pm$13.33\\

{ZScribbleSeg}&\underline{.900$\pm$.065}&\underline{.825$\pm$.069}&.862$\pm$.102&\multicolumn{1}{c|}{.862$\pm$.086}&7.69$\pm$6.94&8.93$\pm$6.40&12.74$\pm$12.48&9.79$\pm$9.19\\
\hdashline
\multicolumn{1}{c|}{FullSupUNet~\citep{baumgartner2017exploration}}&.882$\pm$.123&.824$\pm$.099&.856$\pm$.112&\multicolumn{1}{c|}{.854$\pm$.113}&11.94$\pm$13.58&12.65$\pm$12.52&14.82$\pm$9.69&13.14$\pm$11.97\\
\multicolumn{1}{c|}{\textcolor{black}{FullSup-nnUNet~\citep{isensee2021nnu}}}&\textcolor{black}{.903$\pm$.082}&\textcolor{black}{.817$\pm$.097}&\textcolor{black}{.902$\pm$.063}&\multicolumn{1}{c|}{\textcolor{black}{.874$\pm$.091}}&\textcolor{black}{7.71$\pm$9.83}&\textcolor{black}{7.81$\pm$7.61}&\textcolor{black}{5.15$\pm$2.64}&\textcolor{black}{6.89$\pm$7.36}\\
\hline
\end{tabular}}
\end{table*}

\begin{figure*}[!t]
\centering
\includegraphics[width = \textwidth]{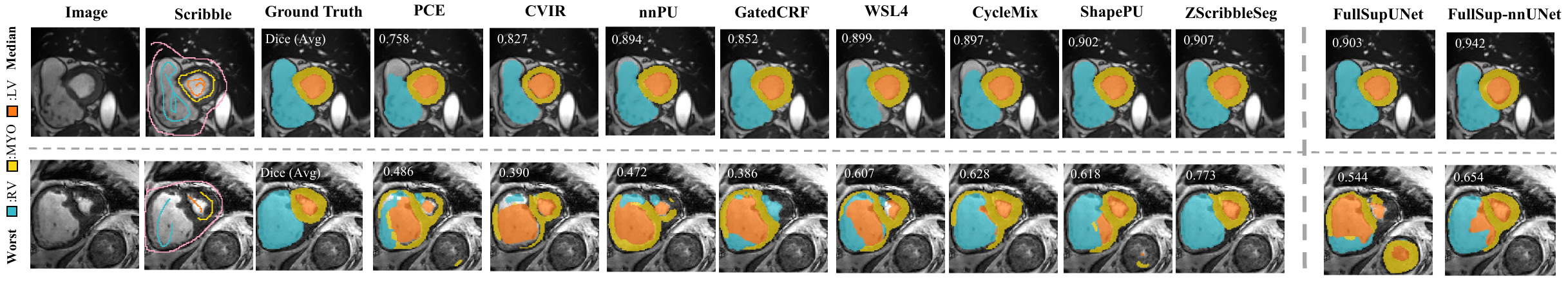} 
\caption{Visualization of cardiac segmentation on ACDC dataset. 
The two slices were from the median and the worst cases by the average Dice scores of all compared methods.}
\label{fig:ACDC-sota}
\end{figure*}
\textbf{Number of annotated points:} 
By varying the number of annotated pixels ($N_{\text{pix}}$), we validated the influence of annotated proportions on scribble-supervised segmentation.
\zxhreffig{fig:exp:scribbleform} (a) and (c) show that the model performance improves as $N_{\text{pix}}$ increases, indicating that model training benefits from a larger proportion of annotated pixels.
One can observe from \zxhreffig{fig:exp:scribbleform} (a) that the segmentation performance started converging when $N_{\text{pix}}$ reached $2n$. For the more difficult segmentation task on irregular objects, the performance converged after    $N_{\text{pix}}\ge 4n$. This trend is illustrated in \zxhreffig{fig:exp:scribbleform} (c).

\textbf{Wide-ranged distribution:}  
We further investigated the influence of a wide range distribution of scribbles with varying step length $l$ in Random walk.
As the step length increases, the label distribution range of Random walk gradually expanded.
From \zxhreffig{fig:exp:scribbleform} (b) and (d), the average Dice score increased as the step length grew, and the performance gradually converged to that of DirRandomWalk.
This confirmed that the widely distributed scribbles were better to provide finer supervision under the same number of draws and annotated pixels.

\subsection{Ablation study}\label{sec-exp-ablation}
We studied the effectiveness of the proposed strategies in modeling efficient scribbles and prior regularization for ZScribbleNet.
We used the ACDC dataset and the expert-made scribble annotations released by \cite{9389796}, and evaluated the model performance on the \textit{validation set}.


We consider eight ablated models trained with different settings. The variants progressively enable the baseline scribble loss $L_{\mathrm{pce}}$, efficiency augmentation (Mix and Occlusion), and the global consistency loss $L_{\mathrm{global}}$, the shape regularization loss $L_{\mathrm{shape}}$, and the spatial prior $L_{\mathrm{spatial}}$), where $L_{\mathrm{spatial}}$ consists of mixture-ratio estimation ($\pi$) and a spatial energy term. 
For example, Model~\#7 corresponds to $L_{\mathrm{pce}}+L_{\mathrm{spatial}}$ (i.e., enabling both $\pi$ and the spatial term), while Model~\#8 is the full method (ZScribbleSeg).

\zxhreftb{tab3} presents the results. 
When model~\#3 incorporated the proposed supervision augmentation for efficient scribble modeling (indicated by the Efficiency column), its performance improved over model~\#1. The average Dice score increased from 0.813 to 0.848, and the average HD decreased from 69.58 mm to 65.35 mm.
Specifically, mixed augmentation brings a 0.5\% marginal improvement (0.818 vs. 0.813). The mix operation increases the distribution range and annotation ratios, but may change the shape of target structure. Therefore, it can be difficult for the segmentation model to learn the shape prior, leading to the HD increase in some structures. This problem is alleviated when combined with the global consistency loss, which helps preserve the shape information. In addition, the occlusion strategy increased the average Dice score by 3.0\% (0.848 vs. 0.818) by enhancing annotation diversity.

\begin{table*}[!h]
\caption{Results and comparisons of regular structure segmentation on pathology enhanced images (LGE CMR) using MSCMRseg dataset. Note that the results of HELPNet and ScribFormer are taken from their original papers, and it can be difficult to conduct a fair cross-study comparison due to the differences in training strategies and hardware environments.} \label{tab5}
\centering
\resizebox{\textwidth}{!}{
\begin{tabular}{ccccccccc}
\hline
\multirow{2}{*}[-4pt]{Methods}&\multicolumn{4}{c|}{Dice}&\multicolumn{4}{c}{HD (mm)}\\
\cmidrule(lr){2-5}\cmidrule(lr){6-9}
&LV & MYO & RV &\multicolumn{1}{c|}{Avg} & LV & MYO & RV & Avg \\
\hline
\multicolumn{1}{c|}{PCE}&.514$\pm$.078&.582$\pm$.067&.058$\pm$.023&\multicolumn{1}{c|}{.385$\pm$.243}&259.4$\pm$14.19&228.1$\pm$21.36&257.4$\pm$12.43&248.3$\pm$21.63\\
\multicolumn{1}{c|}{\textcolor{black}{WSL4~\citep{luo2022scribble}}}&\textcolor{black}{.902$\pm$.040}&\textcolor{black}{.815$\pm$.033}&\textcolor{black}{.828$\pm$.101}&\multicolumn{1}{c|}{\textcolor{black}{.848$\pm$.076}}&\textcolor{black}{55.95$\pm$4.88}&\textcolor{black}{42.07$\pm$13.48}&\textcolor{black}{\underline{32.08$\pm$6.57}}&\textcolor{black}{43.37$\pm$31.04}\\
\multicolumn{1}{c|}{GatedCRF~\citep{Obukhov2019GatedCL}}&.917$\pm$.044&.825$\pm$.032&.848$\pm$.073&\multicolumn{1}{c|}{.863$\pm$.066}&{25.72$\pm$4.37}&37.92$\pm$5.10&32.83$\pm$5.59&32.16$\pm$7.11\\
\multicolumn{1}{c|}{CVIR~\citep{garg2021mixture}}&.331$\pm$.076&.371$\pm$.088&.404$\pm$.110&\multicolumn{1}{c|}{.368$\pm$.095}&259.2$\pm$14.23&243.0$\pm$13.76&180.9$\pm$55.44&227.7$\pm$47.63\\
\multicolumn{1}{c|}{nnPU~\citep{NIPS2017_7cce53cf}}&.341$\pm$.067&.538$\pm$.081&.432$\pm$.100&\multicolumn{1}{c|}{.437$\pm$.115}&259.4$\pm$14.19&201.6$\pm$66.98&199.7$\pm$57.50&220.2$\pm$57.70\\		\multicolumn{1}{c|}{CycleMix~\citep{zhang2022cyclemix}}&.748$\pm$.064&.730$\pm$.047&.835$\pm$.041&\multicolumn{1}{c|}{.771$\pm$.069}&224.59$\pm$35.27&\underline{28.26$\pm$20.77}&73.36$\pm$51.39&108.74$\pm$92.65\\ \multicolumn{1}{c|}{ShapePU~\citep{zhang2022shapepu}}&.880$\pm$.046&.785$\pm$.080&.833$\pm$.087&\multicolumn{1}{c|}{.833$\pm$.082}&178.02$\pm$50.93&178.05$\pm$25.39&189.35$\pm$55.78&181.81$\pm$45.27\\
\multicolumn{1}{c|}{TIP25~\citep{chentip25}}&
.901$\pm$.054&
.823$\pm$.087&
\underline{.873$\pm$.091}&
\multicolumn{1}{c|}{.866$\pm$.083}&
\underline{23.33$\pm$26.28}&
25.42$\pm$27.60&
\textbf{30.60$\pm$37.23}&
\textbf{26.45$\pm$23.67}\\
\multicolumn{1}{c|}{HELPNet~\citep{zhangmedia25}}&
\textbf{.932$\pm$.040}&
\textbf{.861$\pm$.040}&
\textbf{.896$\pm$.040}&
\multicolumn{1}{c|}{\textbf{.896}}&
-&
-&
-&
-\\
\multicolumn{1}{c|}{ScribFormer~\citep{li2024scribformer}}&
.896&
.807&
.813&
\multicolumn{1}{c|}{.839}&
-&
-&
-&
-\\
\multicolumn{1}{c|}{ZScribbleSeg}&\underline{.922$\pm$.039}&\underline{.834$\pm$.039}&.854$\pm$.055&\multicolumn{1}{c|}{\underline{.870$\pm$.058}}&\textbf{12.10$\pm$14.70}&\textbf{16.52$\pm$19.14}&51.03$\pm$39.27&\underline{26.55$\pm$31.39}\\
\hdashline
\multicolumn{1}{c|}{FullSupUNet~\citep{baumgartner2017exploration}}&.909$\pm$.049&.821$\pm$.054&.826$\pm$.087&\multicolumn{1}{c|}{.852$\pm$.076}&10.02$\pm$12.36&11.89$\pm$11.34&56.91$\pm$41.99&26.27$\pm$33.63\\
\multicolumn{1}{c|}{\textcolor{black}{FullSup-nnUNet~\citep{isensee2021nnu}}}&\textcolor{black}{.940$\pm$.034}&\textcolor{black}{.880$\pm$.027}&\textcolor{black}{.902$\pm$.047}&\multicolumn{1}{c|}{\textcolor{black}{.907$\pm$.044}}&\textcolor{black}{9.10$\pm$17.10}&\textcolor{black}{10.57$\pm$11.81}&\textcolor{black}{11.36$\pm$11.50}&\textcolor{black}{10.34$\pm$13.76}\\
\hline
\end{tabular}}
\end{table*} 

\begin{figure*}[!thb]
\centering
\includegraphics[width = \textwidth]{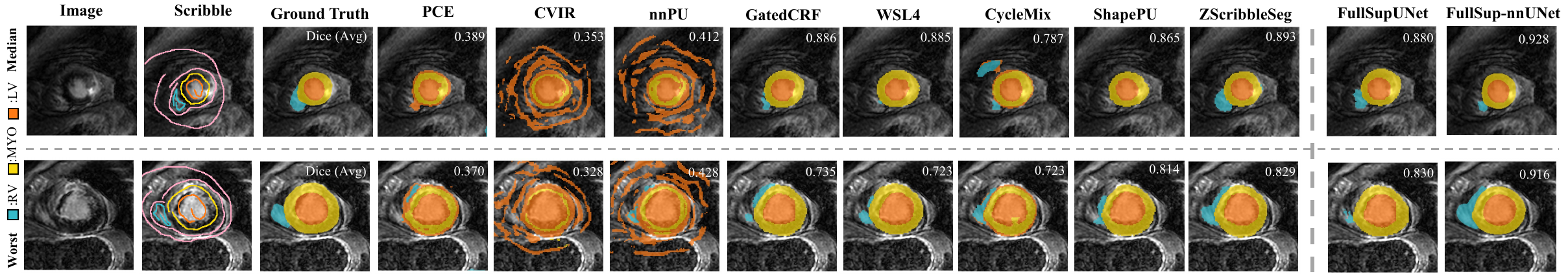}
\caption{Visualization of cardiac segmentation on LGE CMR using MSCMRseg dataset. The two slices were from the median and the worst cases by the average Dice scores of all compared methods.}
\label{fig:mscmr-sota}
\end{figure*}

\begin{table*}[!h]
\caption{Results and comparisons of regular structure segmentation on BTCV dataset.}
\label{tab6}
\centering
\resizebox{\textwidth}{!}{
{
\begin{tabular}{ccccccccc}
\hline
\multirow{2}{*}[-4pt]{Methods}&\multicolumn{4}{c|}{Dice}&\multicolumn{4}{c}{HD (mm)}\\
\cmidrule(lr){2-5}\cmidrule(lr){6-9}
&Spleen & Liver & Stomach &\multicolumn{1}{c|}{Avg} & Spleen & Liver & Stomach & Avg \\
\hline
\multicolumn{1}{c|}{PCE} 
& .384$\pm$.091 & .855$\pm$.021 & .725$\pm$.081 & \multicolumn{1}{c|}{.655$\pm$.226}
& 213.93$\pm$17.98 & 68.61$\pm$19.89 & 100.68$\pm$31.85 & 127.74$\pm$15.18 \\

\multicolumn{1}{c|}{WSL4~\citep{luo2022scribble}} 
& .584$\pm$.091 & .920$\pm$.015 & .818$\pm$.059 & \multicolumn{1}{c|}{.774$\pm$.037}
& 115.58$\pm$19.81 & 31.50$\pm$10.33 & 49.64$\pm$24.89 & 65.67$\pm$12.01  \\

\multicolumn{1}{c|}{GatedCRF~\citep{obukhov2019gated}} 
& .608$\pm$.095 & .919$\pm$.021 & .825$\pm$.060 & \multicolumn{1}{c|}{.783$\pm$.042}
& 109.10$\pm$20.14 & 44.97$\pm$20.93 & 35.49$\pm$39.91 & 63.21$\pm$14.23 \\

\multicolumn{1}{c|}{CVIR~\citep{garg2021mixture}} 
& .596$\pm$.072 & .829$\pm$.046 & .724$\pm$.099 & \multicolumn{1}{c|}{.716$\pm$.058}
& 112.96$\pm$36.4 & 77.88$\pm$25.96 & 102.56$\pm$30.04 & 97.80$\pm$26.05 \\

\multicolumn{1}{c|}{nnPU~\citep{NIPS2017_7cce53cf}} 
& .557$\pm$.101 & .915$\pm$.027 & .809$\pm$.063 & \multicolumn{1}{c|}{.760$\pm$.045}
& 128.65$\pm$19.25 & 34.76$\pm$15.05 & 75.59$\pm$21.45 & 79.67$\pm$18.56 \\
\multicolumn{1}{c|}{CycleMix~\citep{zhang2022cyclemix}}&.738$\pm$.074&.922$\pm$.097&.823$\pm$.091&\multicolumn{1}{c|}{.827$\pm$.095}&89.09$\pm$16.72&29.35$\pm$12.88&32.52$\pm$21.39&50.32$\pm$19.63\\
\multicolumn{1}{c|}{ShapePU~\cite{zhang2022shapepu}}&.745$\pm$.092&.926$\pm$.067&.819$\pm$.076&\multicolumn{1}{c|}{.830$\pm$.081}&76.09$\pm$25.45&25.29$\pm$10.18&48.43$\pm$24.23&49.94$\pm$23.12\\
\multicolumn{1}{c|}{TIP25~\citep{chentip25}}&.780$\pm$.066&.932$\pm$.080&.837$\pm$.075&\multicolumn{1}{c|}{.850$\pm$.073}&49.88$\pm$10.73&20.15$\pm$13.28&22.82$\pm$15.31&30.95$\pm$14.22\\
\multicolumn{1}{c|}{HELPNet~\citep{zhangmedia25}}&\textbf{.796$\pm$.045}&.940$\pm$.088&\textbf{.842$\pm$.090}&\multicolumn{1}{c|}{\textbf{.859$\pm$.075}}&\textbf{36.48$\pm$9.43}&\underline{18.97$\pm$5.34}&\underline{21.31$\pm$8.27}&\underline{25.59$\pm$8.42}\\
\multicolumn{1}{c|}{ScribFormer~\citep{li2024scribformer}}&.752$\pm$.175&.930$\pm$.212&.821$\pm$.370&\multicolumn{1}{c|}{.834$\pm$.379}&62.54$\pm$13.87&21.97$\pm$10.34&33.38$\pm$15.27&39.30$\pm$14.57\\
\multicolumn{1}{c|}{ZScribbleSeg}&\underline{.784$\pm$.034}&\textbf{.947$\pm$.012}&\underline{.839$\pm$.066}&\multicolumn{1}{c|}{\underline{.856$\pm$.080}}&\underline{38.39.$\pm$4.30}&\textbf{12.17$\pm$6.83}&\textbf{17.72$\pm$5.35}&\textbf{22.76$\pm$6.32}\\
\hdashline
\multicolumn{1}{c|}{FullSupUNet~\citep{baumgartner2017exploration}}&.825$\pm$.046&.961$\pm$.032&.867$\pm$.044&\multicolumn{1}{c|}{.884$\pm$.076}&23.72$\pm$5.44&4.31$\pm$1.33&10.81$\pm$6.46&12.95$\pm$6.10\\
\multicolumn{1}{c|}
{FullSup-nnUNet~\citep{isensee2021nnu}}&.838$\pm$.039&.968$\pm$.045&.875$\pm$.069&\multicolumn{1}{c|}{.893$\pm$.062}&19.46$\pm$4.89&4.24$\pm$0.98&10.25$\pm$4.46&11.31$\pm$5.19\\
\hline
\end{tabular}
}}
\end{table*}

When supervision augmentation was combined with the global consistency loss ($\mathcal{L}_{\text{global}}$) in model~\#4, the performance was further enhanced. The average Dice score increased by 4.3\% (0.891 vs. 0.848), and the average HD was reduced by more than 45 mm (19.50 mm vs. 65.35 mm). 
Alternatively, when inter-connectivity was enforced through the shape regularization loss ($\mathcal{L}_{\text{shape}}$), model~\#5 achieved a dramatic reduction in HD, from 69.58 mm to 15.16 mm compared with model~\#1. 
We next examined the effect of incorporating the spatial prior ($\mathcal{L}_{\text{spatial}}$) into ZScribbleNet. Model~\#7, which added this single loss, achieved the largest Dice improvement of 8.1\% (0.894 vs. 0.813). In particular, when only the estimated class mixture ratio $\bm{\pi}$ was applied, model~\#6 still obtained a notable gain of 6.8
When combined with the computed spatial energy, model~\#7 further improved the Dice score from 88.1\% to 89.4\%, and reduced the average HD from 44.25 mm to 27.08 mm, demonstrating the effectiveness of incorporating the spatial prior.
Finally, our ZScribbleSeg (model~\#8) achieved the best performance with an average Dice of 0.899 and HD of $8.70$ mm.
This indicated that the combination of efficient scribbles and priors endowed the algorithm with substantial supervision and prior knowledge for scribble-supervised segmentation.

\subsection{Performance and Comparisons} 
\label{sec-exp-sota}
    
We conducted experiments over the six segmentation tasks stated in Section 4.1.
\textbf{(1)} For the structural segmentation of cardiac ventricles from ACDC dataset, we used the expert-made scribbles released by \cite{9389796}.
\textbf{(2)} For the cardiac structural segmentation from pathology enhanced imaging (MSCMRseg) dataset,  we used the manually annotated scribbles  released by~\cite{zhang2022cyclemix}. 
\textbf{(3)} For the irregular myocardial pathology segmentation from MyoPS dataset, we first adopted the standard skeletonization algorithm for the simulated scribble annotation of pathologies \cite{rajchl2017employing}. Then, we manually annotated skeleton scribbles for the structures of LV, Myo, RV and background.
\textbf{(4)} For the 3D Prostate-Decathlon dataset, we randomly select about 3-5 slices for each category in each dimension to manually label scribbles. 
\textbf{(5)} For Decathlon-BrainTumor and BTCV, we generated pseudo-scribbles on axial slices using a script designed to imitate human drawing behavior~\citep{luo2022scribble}.

\begin{table*}[!h]
\caption{Results and comparisons of irregular segmentation of myocardial pathologies on MyoPS dataset.}\label{tab7}
\centering
\resizebox{0.9\textwidth}{!}{
\begin{tabular}{ccccccc}
\hline
\multirow{2}{*}[-4pt]{Methods}&\multicolumn{3}{c|}{Dice}&\multicolumn{3}{c}{HD (mm)}\\
\cmidrule(lr){2-4}\cmidrule(lr){5-7}
& Scar & Edema &\multicolumn{1}{c|}{Avg} & Scar & Edema & Avg \\
\hline
\multicolumn{1}{c|}{PCE}&0.504$\pm$0.213&0.057$\pm$0.022&\multicolumn{1}{c|}{0.281$\pm$0.271}&82.68$\pm$33.95&147.61$\pm$20.59&115.15$\pm$43.00\\
\multicolumn{1}{c|}{WSL4~\citep{luo2022scribble}}&-&-&\multicolumn{1}{c|}{-}&-&-&-\\
\multicolumn{1}{c|}{GatedCRF~\citep{obukhov2019gated}}&-&-&\multicolumn{1}{c|}{-}&-&-&-\\
\multicolumn{1}{c|}{CVIR~\citep{garg2021mixture}}&0.505$\pm$0.214&0.080$\pm$0.031&\multicolumn{1}{c|}{0.293$\pm$0.263}&61.59$\pm$32.09&125.27$\pm$20.83&93.43$\pm$41.86\\	\multicolumn{1}{c|}{nnPU~\citep{NIPS2017_68053af2}}&0.530$\pm$0.241&0.085$\pm$0.035&\multicolumn{1}{c|}{0.308$\pm$0.282}&48.88$\pm$23.55&125.27$\pm$20.83&87.07$\pm$44.47\\		\multicolumn{1}{c|}{CycleMix~\citep{zhang2022cyclemix}}&0.550$\pm$0.237&0.626$\pm$0.124&\multicolumn{1}{c|}{0.588$\pm$0.191}&65.64$\pm$42.81&81.97$\pm$40.87&73.81$\pm$42.13\\
\multicolumn{1}{c|}{ShapePU~\citep{zhang2022shapepu}}&0.558$\pm$0.237&0.615$\pm$0.144&\multicolumn{1}{c|}{0.587$\pm$0.205}&57.33$\pm$31.58&53.00$\pm$31.42&55.16$\pm$31.17\\
\multicolumn{1}{c|}{TIP25~\citep{chentip25}}&
\underline{0.603$\pm$0.175}&
0.652$\pm$0.141&
\multicolumn{1}{c|}{0.628$\pm$0.193}&
49.29$\pm$20.31&
50.23$\pm$22.56&
49.76$\pm$21.44\\
\multicolumn{1}{c|}{HELPNet~\citep{zhangmedia25}}&
\textbf{0.611$\pm$0.204}&
\underline{0.655$\pm$0.147}&
\multicolumn{1}{c|}{\underline{0.633$\pm$0.186}}&
\textbf{45.89$\pm$20.12}&
\textbf{47.74$\pm$21.34}&
\textbf{46.82$\pm$20.89}\\
\multicolumn{1}{c|}{ScribFormer~\citep{li2024scribformer}}&
0.571$\pm$0.266&
0.648$\pm$0.190&
\multicolumn{1}{c|}{0.610$\pm$0.228}&
50.55$\pm$19.09&
51.28$\pm$24.21&
50.92$\pm$21.65\\
\multicolumn{1}{c|}{ZScribbleSeg}&0.596$\pm$0.237&\textbf{0.676$\pm$0.113}&\multicolumn{1}{c|}{\textbf{0.636$\pm$0.188}}&\underline{46.73$\pm$20.04}&\underline{47.05$\pm$24.30}&\underline{46.89$\pm$21.98}\\
\hdashline
\multicolumn{1}{c|}{FullSupUNet~\citep{baumgartner2017exploration}}&0.607$\pm$0.253&0.659$\pm$0.135&\multicolumn{1}{c|}{0.633$\pm$0.202}&55.35$\pm$35.73&63.53$\pm$33.15&59.44$\pm$34.27\\
\multicolumn{1}{c|}
{\textcolor{black}{FullSup-nnUNet~\citep{isensee2021nnu}}}&\textcolor{black}{0.610$\pm$0.169}&\textcolor{black}{0.651$\pm$0.246}&\multicolumn{1}{c|}{\textcolor{black}{0.630$\pm$0.209}}&\textcolor{black}{33.89$\pm$14.00}&\textcolor{black}{32.55$\pm$17.88}&\textcolor{black}{33.22$\pm$15.87}\\
\hline
\end{tabular}}
\end{table*}

\begin{figure*}[!thb]
\centering
\includegraphics[width = 1\textwidth]{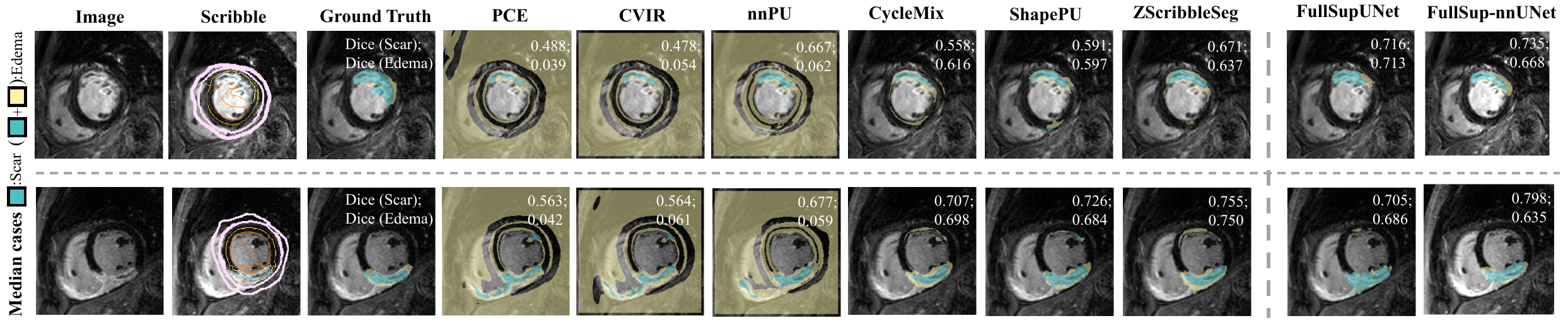}
\caption{Visualization of irregular segmentation of myocardial pathologies on MyoPS dataset. The two slices were from the median cases by average Dice scores of edema or scar segmentation of all compared methods.
}
\label{fig:myops-sota}
\end{figure*}

\begin{table*}[!h]
\caption{Results and comparisons of irregular structure segmentation using Decathlon-BrainTumor dataset.}
\label{tab8}
\centering
\resizebox{0.9\textwidth}{!}{
{
\begin{tabular}{ccccccc}
\hline
\multirow{2}{*}[-4pt]{Methods}&\multicolumn{3}{c|}{Dice}&\multicolumn{3}{c}{HD (mm)}\\
\cmidrule(lr){2-4}\cmidrule(lr){5-7}
& Tumor & Edema &\multicolumn{1}{c|}{Avg} & Tumor & Edema & Avg \\
\hline
\multicolumn{1}{c|}{PCE}
&0.683$\pm$0.215 & 0.569$\pm$0.312 & \multicolumn{1}{c|}{0.626$\pm$0.288}
&89.84$\pm$42.16 & 68.13$\pm$31.54 & 78.99$\pm$38.86 \\

\multicolumn{1}{c|}{WSL4~\citep{luo2022scribble}}
& 0.709$\pm$0.235 & 0.669$\pm$0.167 & \multicolumn{1}{c|}{0.689$\pm$0.188}
& 76.29$\pm$31.24 & 63.28$\pm$27.23 & 69.79$\pm$28.34 \\

\multicolumn{1}{c|}{GatedCRF~\citep{Obukhov2019GatedCL}}
& 0.713$\pm$0.244 & 0.670$\pm$0.297 & \multicolumn{1}{c|}{0.692$\pm$0.284 }
& 74.38$\pm$29.24 & 60.51$\pm$21.60 & 67.45$\pm$25.36 \\

\multicolumn{1}{c|}{CVIR~\citep{garg2021mixture}}
& 0.720$\pm$0.265 & 0.667$\pm$0.343 & \multicolumn{1}{c|}{0.694$\pm$0.279}
& 71.15$\pm$38.43 & 59.42$\pm$24.30 & 65.29$\pm$28.41 \\

\multicolumn{1}{c|}{nnPU~\citep{NIPS2017_7cce53cf}}
& 0.708$\pm$0.350 & 0.648$\pm$0.389 & \multicolumn{1}{c|}{0.678$\pm$0.364}
& 80.49$\pm$34.27 & 66.29$\pm$26.70 & 73.39$\pm$29.16 \\

\multicolumn{1}{c|}{CycleMix~\citep{zhang2022cyclemix}}
&0.741$\pm$0.389&0.711$\pm$0.236&\multicolumn{1}{c|}{0.726$\pm$0.253}
&38.56$\pm$25.82&37.55$\pm$21.57&38.10$\pm$23.88\\

\multicolumn{1}{c|}{ShapePU~\citep{zhang2022shapepu}}
&0.750$\pm$0.296&0.712$\pm$0.236&\multicolumn{1}{c|}{0.731$\pm$0.272}
&33.29$\pm$19.42&30.16$\pm$17.57&31.73$\pm$18.13\\

\multicolumn{1}{c|}{TIP25~\citep{chentip25}}
&0.784$\pm$0.077&0.724$\pm$0.078&\multicolumn{1}{c|}{0.754$\pm$0.081}
&\underline{21.82}$\pm$13.94&21.88$\pm$10.45&\underline{21.85$\pm$12.64}\\

\multicolumn{1}{c|}{HELPNet~\citep{zhangmedia25}}
&\textbf{0.811$\pm$0.054}&\underline{0.734$\pm$0.086}&\multicolumn{1}{c|}{\textbf{0.772$\pm$0.077}}
&\textbf{20.69$\pm$12.66}&\textbf{19.57$\pm$9.10}&\textbf{20.13$\pm$10.98}\\

\multicolumn{1}{c|}{ScribFormer~\citep{li2024scribformer}}
&0.773$\pm$0.086&0.720$\pm$0.109&\multicolumn{1}{c|}{0.747$\pm$0.090}
&27.28$\pm$9.38&24.81$\pm$14.83&26.05$\pm$13.22\\

\multicolumn{1}{c|}{ZScribbleSeg}
&\underline{0.788$\pm$0.089}&\textbf{0.737$\pm$0.098}&\multicolumn{1}{c|}{\underline{0.763$\pm$0.098}}
&24.28$\pm$12.13&\underline{19.91$\pm$8.14}&22.10$\pm$10.67\\

\hdashline
\multicolumn{1}{c|}{FullSupUNet~\citep{baumgartner2017exploration}}
&0.803$\pm$0.097&0.821$\pm$0.156&\multicolumn{1}{c|}{0.812$\pm$0.112}
&21.34$\pm$4.87&12.38$\pm$5.63&16.86$\pm$5.38\\
\multicolumn{1}{c|}{FullSup-nnUNet~\citep{isensee2021nnu}}
&0.816$\pm$0.078&0.829$\pm$0.218&\multicolumn{1}{c|}{0.823$\pm$0.191}
&18.87$\pm$5.69&11.01$\pm$4.67&14.94$\pm$4.97\\
\hline
\end{tabular}
}}
\end{table*}
\hspace*{\fill}

We compared ZScribbleSeg with thirteen methods.
We first implemented the PCE loss ($\mathcal{L}_{\text{pce}}$) as a baseline method (referred to PCE).
We reported results of state-of-the-art scribble-supervised methods, including HELPNet~\citep{zhangmedia25}, TIP25~\citep{chentip25}, ScribFormer~\citep{li2024scribformer}, WSL4~\citep{luo2022scribble}, GatedCRF~\citep{Obukhov2019GatedCL}, CycleMix~\citep{zhang2022cyclemix}, and ShapePU~\citep{zhang2022shapepu}. MAAG~\citep{9389796} was included on ACDC with results taken from the original paper. Note that for the MSCMRSeg dataset, the test sets used by HELPNet and ScribFormer are the same as ours. Therefore, we also report the results of their original papers.
Furthermore, we considered semi-supervised methods based on positive–unlabeled learning, namely CVIR~\citep{garg2021mixture} and nnPU~\citep{NIPS2017_7cce53cf}. We re-implemented both for scribble-supervised segmentation.
Finally, we trained UNet and nnUNet with full annotations as the baselines of fully-supervised approach (referred to as FullSupUNet and FullSup-nnUNet, respectively). For the details of compared methods, please refer to the Section 2 of the supplementary materials.

\subsubsection{Structure segmentation from anatomical images} \label{se-exp-sota:1}
\zxhreftb{tab5} presents the Dice and HD results of 11 approaches for regular structure segmentation of cardiac ventricles from ACDC dataset. 
One can observe that ZScribbleSeg achieved an average Dice of 0.862, while TIP25, HELPNet, and ScribFormer obtained higher Dice scores of 0.863, 0.901, and 0.856, respectively. We further evaluate regular abdominal organ segmentation on the BTCV dataset. As \zxhreftb{tab6} shows, among all compared methods, HELPNet obtained highest average Dice (0.859) and spleen HD, while ZScribbleSeg achieved an average Dice of 0.856 and the lowest average HD of 22.76\,mm. 
Particularly, the HD results of ZScribbleSeg were better than the other methods except on the spleen. 
Note that HD is highly sensitive to the noisy and outlier segmentation results, which are commonly seen when the supervision of global shape information is not sufficient. 
The results indicate the proposed efficient scribble modeling and prior regularization were able to alleviate the problem of inadequate supervision and incomplete shape information from training images with scribble annotations.
Finally, \zxhreffig{fig:ACDC-sota} visualizes two typical cases (median and worst) of ACDC dataset for illustration.

\subsubsection{Structure segmentation from pathology enhanced images} 
The anatomical segmentation from pathology-enhanced images was more challenging than that of the ACDC dataset. This is mainly because MSCMRseg provides fewer training subjects (25 vs. 70) and the LGE CMR images exhibit lower quality and more complex appearance patterns.

\zxhreftb{tab5} provides the quantitative results, and \zxhreffig{fig:mscmr-sota} visualizes two special examples (median and worst) for demonstration.
ZScribbleSeg achieved promising performance with an average Dice of 0.870 and HD of 26.55 mm, which are the second best performance among all compared method.
Notice that for this particular challenging task, the two general semi-supervised methods, CVIR and nnPU, failed to work properly, as evidenced by the unsuccessful segmentation examples shown in \zxhreffig{fig:mscmr-sota}.

Finally, similar to the results in previous study (Section~\ref{se-exp-sota:1}), ZScribbleSeg and FullSupUNet could achieve less noisy segmentation. This was confirmed by their substantially better HD results in \zxhreftb{tab6}.
Hence, we conclude that ZScribbleNet benefits from substantially augmented supervision and global shape information through the proposed efficient scribble modeling and prior regularization.

\begin{table*}[!h]
\caption{Results and comparisons of \emph{3D segmentation on Decatholon Prostate dataset} with 5-fold cross validation. The results were reported on the test set.}\label{tab1_6_v0}
\centering
\resizebox{\textwidth}{!}{
\begin{tabular}{ccccccc}
\hline
\multirow{2}{*}[-4pt]{Methods}&\multicolumn{3}{c|}{Dice}&\multicolumn{3}{c}{HD (mm)}\\
\cmidrule(lr){2-4}\cmidrule(lr){5-7}
&Central gland & Peripheral Zone &\multicolumn{1}{c|}{Avg} & Central gland & Peripheral Zone & Avg \\
\hline
\multicolumn{1}{c|}{PCE}&0.784$\pm$0.138&0.535$\pm$0.190&\multicolumn{1}{c|}{0.659$\pm$0.213}&19.85$\pm$15.58&25.73$\pm$12.72&22.79$\pm$15.31\\
\multicolumn{1}{c|}{WSL4~\citep{luo2022scribble}}&-&-&\multicolumn{1}{c|}{-}&-&-&-\\
\multicolumn{1}{c|}
{GatedCRF~\cite{obukhov2019gated}}&0.776$\pm$0.112&0.555$\pm$0.177&\multicolumn{1}{c|}{0.666$\pm$0.123}&31.90$\pm$21.82&31.41$\pm$117.40&31.66$\pm$18.43\\

\multicolumn{1}{c|}{CVIR~\citep{garg2021mixture}}&0.780$\pm$0.139&0.518$\pm$0.222&\multicolumn{1}{c|}{0.649$\pm$0.238}&\underline{15.26$\pm$10.37}&22.93$\pm$9.94&\textbf{19.10$\pm$11.17}\\	

\multicolumn{1}{c|}
{nnPU~\citep{NIPS2017_68053af2}}&0.779$\pm$0.133&0.529$\pm$0.196&\multicolumn{1}{c|}{0.654$\pm$0.218}&19.29$\pm$12.29&23.49$\pm$10.21&21.39$\pm$11.81\\
\multicolumn{1}{c|}{CycleMix~\citep{zhang2022cyclemix}}&\underline{0.796$\pm$0.159}&0.583$\pm$0.218&\multicolumn{1}{c|}{\underline{0.689$\pm$0.229}}&17.31$\pm$15.48&\textbf{21.44$\pm$11.47}&19.37$\pm$14.01\\
\multicolumn{1}{c|}{ShapePU~\citep{zhang2022shapepu}}&0.792$\pm$0.180&\underline{0.587$\pm$0.243}&\multicolumn{1}{c|}{\underline{0.689$\pm$0.240}}&15.46$\pm$10.87&\underline{23.15$\pm$13.09}&19.31$\pm$12.90\\
\multicolumn{1}{c|}{ZScribbleSeg}&\textbf{0.799$\pm$0.142}&\textbf{0.612$\pm$0.204}&\multicolumn{1}{c|}{\textbf{0.706$\pm$0.207}}&\textbf{14.81$\pm$8.82}&23.76$\pm$18.26&\underline{19.28$\pm$14.93}\\
\hdashline
\multicolumn{1}{c|}{FullSupUNet~\citep{baumgartner2017exploration}}&0.776$\pm$0.138&0.547$\pm$0.185&\multicolumn{1}{c|}{0.661$\pm$0.212}&15.37$\pm$12.19&19.18$\pm$12.38&17.28$\pm$13.14\\
\multicolumn{1}{c|}{FullSup-nnUNet~\citep{isensee2021nnu}}&0.845$\pm$0.095&0.607$\pm$0.252&\multicolumn{1}{c|}{0.726$\pm$0.225}&8.57$\pm$2.77&19.17$\pm$14.25&13.87$\pm$11.30\\
\hline
\end{tabular}}
\end{table*}

\begin{figure*}[!t]
\centering
\includegraphics[width =\textwidth]{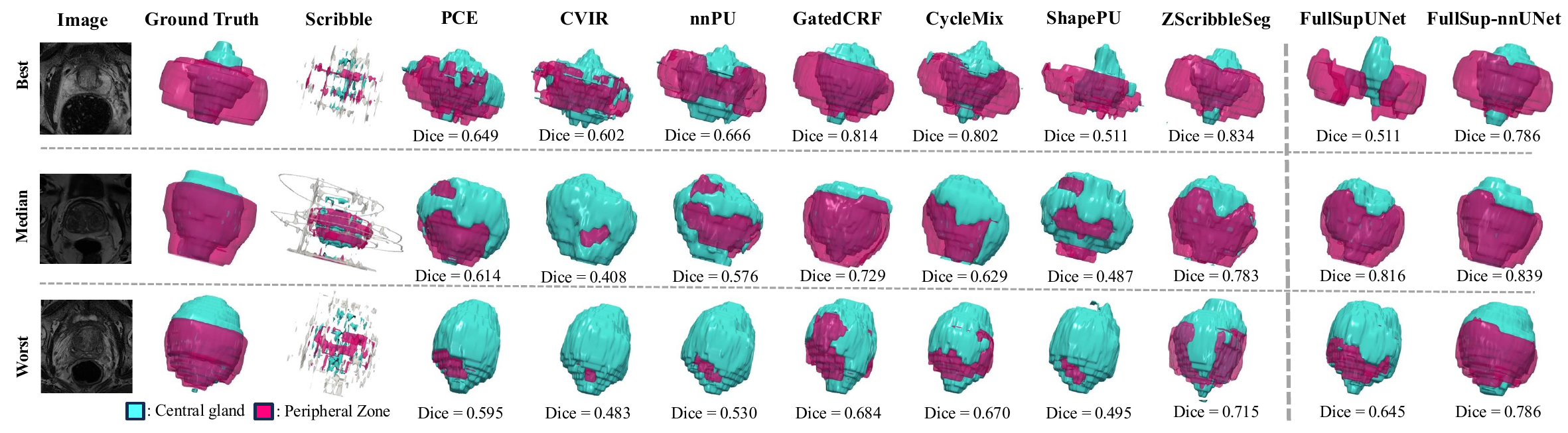}
\caption{Visualization of results on Prostate dataset.  The selected subjects were the best, median and  worst cases by the average Dice scores of all compared methods.}
\label{fig:Prostate-sota}
\end{figure*}

\subsubsection{Irregular pathology segmentation}
For segmentation of objects with heterogeneous shape features, it becomes particularly challenging to learn the accurate shape information for inference. 
We evaluated ZScribbleSeg on such challenging task of irregular segmentation using myocardial pathology segmentation (MyoPS) and Decathlon-BrainTumor datasets, where \textit{we removed the shape regularization term $\mathcal{L}_{\text{shape}}$ due to the nature of pathologies lacking such property}.

\zxhreftb{tab7} presents the detailed performance results. \zxhreffig{fig:myops-sota} further illustrates two representative cases, namely the median examples in terms of average Dice scores for edema and scar segmentation.
One can find that the advantages of the proposed methodologies were demonstrated in such challenging task. ZScribbleSeg achieved the highest average Dice (0.636), while HELPNet and TIP25 also obtained promising performance with average Dice 0.633 and 0.628, respectively. The performance gains of ZScribbleSeg, either in terms of Dice or HD, were significant from CycleMix, ShapePU and finally to ZScribbleSeg compared to PCE, WSL4, GatedCRF, CVIR and nnPU ($p\!\!<\!\!0.001$).
In fact, the five compared methods failed to segment edema, and both WSL4 and GatedCRF also failed on scar segmentation. This is illustrated in the visualized examples in \zxhreffig{fig:myops-sota}. 
Although WSL4 and GatedCRF performed well with scribble supervision in the two regular structure segmentation tasks, they suffered severely from noisy labels in this setting. Their reliance on pseudo labels made the training unstable and led to model failure.
The difficulty is exacerbated by the similar texture of edema and surrounding tissues across imaging modalities. Without robust estimation and regularization of class mixture ratios, accurate delineation becomes infeasible, leading to the universal failure of the compared methods.
By contrast, ZScribbleSeg succeeded in this task thanks to their own methods of estimating the class prior $\bm{\pi}$ and applying spatial regularization. This is confirmed by their averaged HD results on scar and edema, with ShapePU at 55.16 mm and ZScribbleSeg further reduced to 46.89 mm.
Table~\ref{tab8} reports the quantitative results for tumor core and peritumoral edema segmentation. ZScribbleSeg achieved second best performance in terms of Dice, with an average Dice of 0.763. These results further support the robustness of the proposed framework on more challenging irregular pathology segmentation tasks.


\subsubsection{3D Segmentation}
We generalized ZScribbleSeg to 3D cases and validated the proposed framework on the 3D Prostate segmentation (Decathlon-Prostate) sequences . \zxhreftb{tab1_6_v0} summarizes the Dice and HD results of 9 methods. Since the released code of WSL4~\citep{luo2022scribble} is designed for 2D images, we did not include it for this study. 
The proposed ZScribbleSeg obtained the average Dice score of 0.726 and HD of 19.28 mm, surpassing the result of other scribble-supervised methods. We provided detailed results of compared methods on each fold in the Section 3 of the supplementary material. 
\newline\indent Figure~\ref{fig:Prostate-sota} presents three typical cases, including the best, the median, and the worst results selected by the average dice score of all compared methods. ZScribbleSeg generated segmentations with more realistic shape and size than the other scribble based methods, and this advantage was particularly evident for peripheral zone segmentation.

\subsubsection{Scalability and Parameter Sensitivity Analysis}

\begin{figure}[t]
    \centering
    \includegraphics[width=\linewidth]{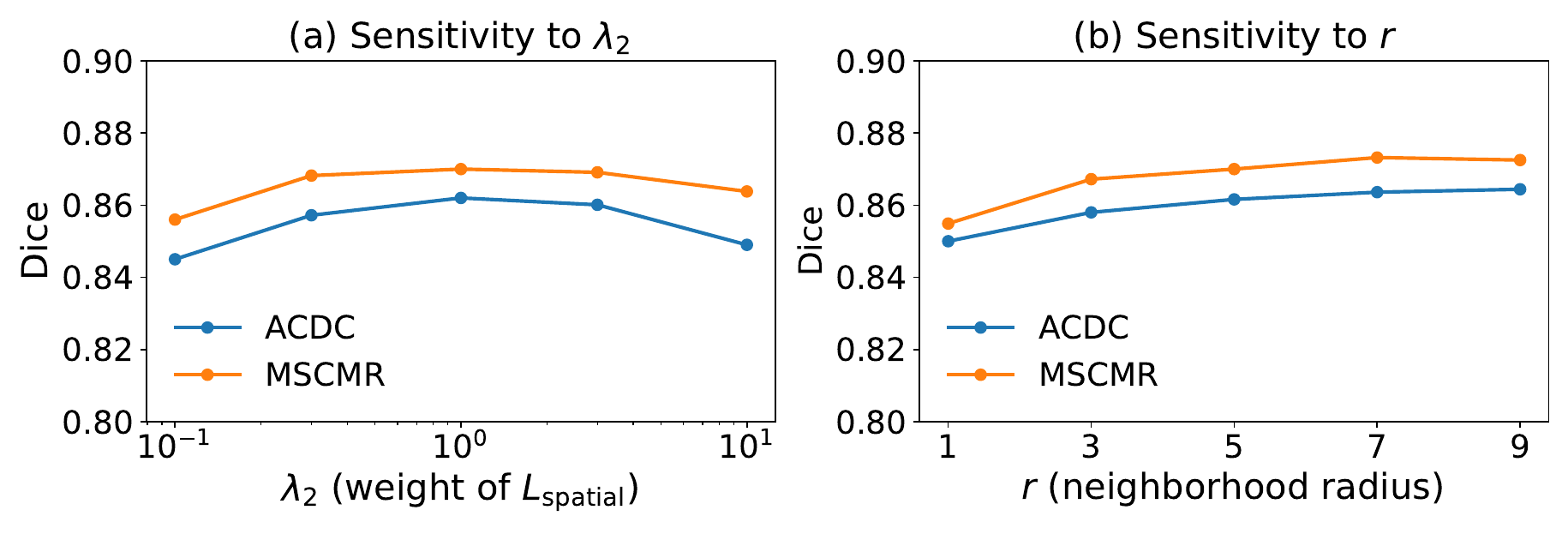}
    \caption{{Sensitivity analysis of key hyperparameters on ACDC and MSCMRseg. We vary the spatial prior weight $\lambda_2$ and the neighborhood radius $r$, while keeping all other settings fixed.}}
    \label{fig:sensitivity}
\end{figure}

Table~\ref{tab:runtime_memory_lspatial} reports the training and inference time together with the peak GPU memory usage with and without $\mathcal{L}_{\text{spatial}}$ on two 2D datasets (ACDC and MyoPS) and one 3D dataset (Prostate). Enabling $\mathcal{L}_{\text{spatial}}$ introduces a moderate training overhead, as it requires computing the spatial energy $\Phi$ and performing local ranking and top-$\pi$ selection. Specifically, the training time increases from 0.1423 to 0.1689 s/iter on ACDC, from 0.1618 to 0.1935 s/iter on MyoPS, and from 0.4823 to 0.5202 s/iter on Prostate. The corresponding increase in peak training memory is marginal. In contrast, inference time and memory remain unchanged across all datasets, since the $\Phi$-based ranking is only used during training and the test-time forward pass is identical to the baseline model. 
Although $\Phi$ can be expressed in a dense pairwise form, in practice it is computed locally within a radius-$r$ neighborhood  (Eq.~\ref{3.3eq11}), rather than over all pixel pairs. As a result, the computational complexity scales as $\mathcal{O}(N \cdot r^2)$ in 2D (and $\mathcal{O}(N \cdot r^3)$ in 3D), where $N$ denotes the number of pixels and $r$ the neighborhood radius. Increasing $r$ enlarges the local window and thus increases the constant factor of the computation quadratically, but does not change the linear asymptotic complexity with respect to image size. 
Figure~\ref{fig:sensitivity} further illustrates the sensitivity to the spatial prior weight $\lambda_2$ and the neighborhood radius $r$ on ACDC and MSCMRseg. The performance is stable across a broad range of $\lambda_2$ values. Increasing $r$ improves Dice scores initially and then saturates.

\begin{table}[t]
\centering
\caption{{Runtime and peak GPU memory with/without $L_{\text{spatial}}$ on ACDC (2D), MyoPS (2D) and Prostate (3D).
Inference time is reported per volume for per 2D image and 3D volume.}}
\label{tab:runtime_memory_lspatial}
\setlength{\tabcolsep}{4pt}
\renewcommand{\arraystretch}{1.08}
\resizebox{\linewidth}{!}{
\begin{tabular}{lcccccccc}
\toprule
\multirow{2}{*}{Dataset} &
\multirow{2}{*}{Dim.} &
\multirow{2}{*}{Input} &
\multirow{2}{*}{Patch size} &
\multirow{2}{*}{$L_{\text{spatial}}$} &
\multicolumn{2}{c}{Train} &
\multicolumn{2}{c}{Inference} \\
\cmidrule(lr){6-7}\cmidrule(lr){8-9}
&&&& & Time (s/iter) & Mem (GB) & Time (s) & Mem (GB) \\
\midrule

\multirow{2}{*}{ACDC}
& \multirow{2}{*}{2D}
& \multirow{2}{*}{1-ch}
& \multirow{2}{*}{$212\times212$}
& w/o  & 0.1423 & 2.658 & 0.0061 & 2.434 \\
&&&& with & 0.1689 & 2.671 & 0.0061 & 2.434 \\
\midrule

\multirow{2}{*}{MyoPS}
& \multirow{2}{*}{2D}
& \multirow{2}{*}{3-ch}
& \multirow{2}{*}{$192\times192$}
& w/o  & 0.1618 & 3.410 & 0.0073 & 2.473 \\
&&&& with & 0.1935 & 3.421 & 0.0073 & 2.473 \\
\midrule

\multirow{2}{*}{Prostate}
& \multirow{2}{*}{3D}
& \multirow{2}{*}{1-ch}
& \multirow{2}{*}{$64\times64\times64$}
& w/o  & 0.4823 & 6.850 & 0.1344 & 4.850 \\
&&&& with & 0.5202 & 6.904 & 0.1344 & 4.850 \\

\bottomrule
\end{tabular}
}
\end{table}

\section{Conclusion} \label{section6}

In this work, we proposed ZScribbleSeg, a new framework for scribble-supervised segmentation that integrates efficient scribbles and prior regularization within a deep neural network (ZScribbleNet).
ZScribbleSeg leverages the principles of effective scribble annotations and augments scribble supervision in ZScribbleNet through mixup–occlusion operations and global consistency regularization.
Then, we explored to capture the global information by incorporating the prior information, particularly with proposals of the spatial prior loss. 
This loss was based on the estimated spatial energy and label class mixture proportions $\bm{\pi}$. 
The former provides a new means to identify the probability of unlabeled pixels belonging to each class without  directly using model predictions; 
The latter was developed based on the EM algorithm and was aimed to correct the problematic prediction via the regularization of the spatial prior loss.

To evaluate the performance of ZScribbleSeg, we investigated a variety of segmentation tasks, including regular structural segmentation of cardiac ventricles from anatomical imaging data (using ACDC dataset), regular structural segmentation of pathology enhanced imaging data (MSCMRseg), irregular object segmentation from multi-modality imaging (MyoPS) and 3D prostate segmentation from multi-modality imaging (Decathlon-Prostate).
Compared with existing scribble-supervised approaches, ZScribbleSeg demonstrates superior performance.
With augmented supervision and prior regularization, ZScribbleSeg performed reliably in challenging scenarios. It showed strong generalizability on small training sets (MSCMRseg) and irregular structure segmentation (MyoPS), where other compared methods failed.

Free-form weak annotations include points, bounding boxes, and textual descriptions in addition to scribbles. In this work, we focused solely on scribble supervision. 
In practice, the supervision may take different forms due to the variety of annotations that exist in hospital filing systems.
Moreover, training segmentation networks for rare diseases remains challenging, particularly for irregular pathological objects due to the extremely limited data.
Hence, future work will explore supervision augmentation and prior estimation for weakly supervised segmentation in rare disease scenarios. This will involve combining multiple forms of free-form annotations.

\section*{Acknowledgment}

This work was funded by the National Natural Science Foundation of China (grant No. 62372115) Shanghai Municipal Education Commission-Artificial Intelligence Initiative to Promote Research Paradigm Reform and Empower Disciplinary Advancement Plan (grant no. 24KXZNA13) Shanghai Oriental Elite Talent (YingCai) Program.

\bibliographystyle{model2-names.bst}\biboptions{authoryear}
\bibliography{cas-refs}

\end{document}